\def\paperID{0000}
\def\confName{CVPR}
\def\confYear{2026}

\def\paperTitle{A Unified Foundation Model for All-in-One Multi-Modal Remote Sensing Image Restoration and Fusion with Language Prompting}

\def\authorBlock{
    Yongchuan Cui $^{1,2}$ \qquad
    Peng Liu $^{1,2,}$\thanks{Corresponding author}\footnotemark[1] \\
    $^{1}$ Aerospace Information Research Institute, Chinese Academy of Sciences, Beijing, China \\
    $^{2}$ School of Electronic, Electrical and Communication Engineering, \\ University of Chinese Academy of Sciences, Beijing, China \\
    {\tt\small yongchuancui@gmail.com, liupeng202303@aircas.ac.cn}
}

\newif\ifreview 
\newif\ifarxiv \newcommand{\arxiv}{\arxivtrue}
\newif\ifcamera 
\newif\ifrebuttal 

\arxiv % blind review; use \arxiv for de-anonymized draft or \cameraready for final

\pdfoutput=1
\documentclass[10pt,twocolumn,letterpaper]{article}
\ifreview \usepackage[review]{cvpr} \fi
\ifarxiv \usepackage[pagenumbers]{cvpr} \fi
\ifrebuttal \usepackage[rebuttal]{cvpr} \fi
\ifcamera \usepackage{cvpr} \fi

\usepackage{graphicx}
\graphicspath{{./figs/paper/}}
\usepackage{amsmath}	
\usepackage{amssymb}	
\usepackage{booktabs}
\usepackage{times}
\usepackage{microtype}
\usepackage{epsfig}
\usepackage{caption}
\usepackage{float}
\usepackage{placeins}
\usepackage{colortbl}
\usepackage{stfloats}
\usepackage{enumitem}
\usepackage{tabularx}
\usepackage{xstring}
\usepackage{multirow}
\usepackage{xspace}
\usepackage{url}
\usepackage{subcaption}
\usepackage[hang,flushmargin]{footmisc}

\ifcamera \usepackage[accsupp]{axessibility} \fi

% ...

\ifarxiv  \fi

\newcommand{\R}[1]{{%
    \textbf{%
        \ifstrequal{#1}{1}{\textcolor{red}{R#1}}{%
        \ifstrequal{#1}{2}{\textcolor{blue}{R#1}}{%
        \ifstrequal{#1}{3}{\textcolor{magenta}{R#1}}{%
        \ifstrequal{#1}{4}{\textcolor{teal}{R#1}}{%
                           \textcolor{cyan}{R#1}%
        }}}}%
    }%
}}
\usepackage{xr-hyper}

\makeatletter
\newcommand*{\addFileDependency}[1]{
  \typeout{(#1)}
  \@addtofilelist{#1}
  \IfFileExists{#1}{}{\typeout{No file #1.}}
}

\makeatother
\newcommand*{\myexternaldocument}[1]{
    \externaldocument{#1}
    \addFileDependency{#1.tex}
    \addFileDependency{#1.aux}
}

\definecolor{cvprblue}{rgb}{0.21,0.49,0.74}
\usepackage[pagebackref,breaklinks,colorlinks,allcolors=cvprblue]{hyperref}
\usepackage[capitalize]{cleveref}
\crefname{section}{Sec.}{Secs.}
\crefname{table}{Table}{Tables}
\crefname{figure}{Fig.}{Figs.}

\ifarxiv \crefname{appendix}{App.}{Apps.}
\else \crefname{appendix}{Suppl.}{Suppls.} \fi

\unless\ifarxiv \myexternaldocument{_supplementary} \fi

\begin{document}
%% TITLE
\title{\paperTitle}
\author{\authorBlock}
\maketitle

\begin{abstract}
Remote sensing imagery suffers from clouds, haze, noise, resolution limits, and sensor heterogeneity. Existing restoration and fusion approaches train separate models per degradation type. In this work, we present \textbf{L}anguage-conditioned \textbf{La}rge-scale \textbf{R}emote \textbf{S}ensing restoration model (\textbf{LLaRS}), the first unified foundation model for multi-modal and multi-task remote sensing low-level vision. LLaRS employs Sinkhorn-Knopp optimal transport to align heterogeneous bands into semantically matched slots, routes features through three complementary mixture-of-experts layers (convolutional experts for spatial patterns, channel-mixing experts for spectral fidelity, and attention experts with low-rank adapters for global context), and stabilizes joint training via step-level dynamic weight adjustment. To train LLaRS, we construct LLaRS1M, a million-scale multi-task dataset spanning eleven restoration and enhancement tasks, integrating real paired observations and controlled synthetic degradations with diverse natural language prompts. Experiments show LLaRS consistently outperforms seven competitive models, and parameter-efficient finetuning experiments demonstrate strong transfer capability and adaptation efficiency on unseen data.

\noindent \texttt{Repo}: \url{https://github.com/yc-cui/LLaRS}.
\end{abstract}

\section{Introduction}
\label{sec:intro}

\begin{figure}[t]
    \centering
    \includegraphics[width=\linewidth]{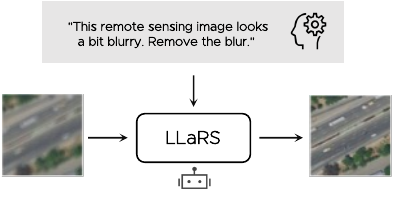}
    \caption{LLaRS takes a degraded remote sensing image and a natural-language prompt as input, and produces the restored image.}
    \label{fig:teaser}
\end{figure}

Remote sensing images support environmental monitoring, resource management, agriculture, and urban planning. However, image acquisition and transmission introduce multiple types of degradation that severely limit their practical utility. These degradations constitute fundamental low-level vision problems: restoring clean, high-quality images from degraded observations~\citep{he2024diffusionmodelslowlevelvision}. Existing approaches to remote sensing image restoration train separate specialist models for each degradation type. This fragmented paradigm scatters research effort across isolated datasets and codebases, ignoring the substantial shared structure among restoration tasks. Each new degradation requires collecting task-specific data, designing custom architectures, and training from scratch, leading to inefficient resource utilization and limited generalization~\cite{Cui_2025_ICCV}.

In contrast, large-scale pretraining and unified modeling have fundamentally reshaped natural language processing and general computer vision, where foundation models learn transferable representations from diverse data and adapt efficiently to downstream tasks. Recent studies document the emergence of geospatial foundation models~\citep{zhou2024visionlanguagegeofoundationmodelsurvey,lu2024aifoundationmodelsremote}, yet remote sensing low-level vision remains underexplored. The field still lacks large-scale unified datasets spanning multiple restoration tasks, architectures capable of processing heterogeneous multi-sensor inputs in a single forward pass, and systematic studies of parameter-efficient adaptation strategies. Bridging this gap requires rethinking the restoration pipeline from a unified, multi-task perspective.

To this end, this paper presents LLaRS (see \cref{fig:teaser}), a language-conditioned~\citep{InstructIR} foundation model that unifies eleven restoration and enhancement tasks under natural-language guidance. Users specify the desired operation through intuitive text prompts such as \textit{remove the blur} or \textit{enhance resolution}, and the model routes the input through task-appropriate processing pathways. We also construct LLaRS1M, a million-scale multi-task dataset integrating real paired observations and controlled synthetic degradations across dehazing, denoising, cloud removal, super-resolution, pansharpening, spatiotemporal fusion, and five additional tasks, each annotated with diverse natural-language prompts. The LLaRS architecture addresses three core technical challenges. First, heterogeneous sensor inputs with varying channel counts and spectral meanings are aligned into semantically matched slots via entropy-regularized optimal transport solved with Sinkhorn iterations~\citep{sinkhorn1967concerning,cuturi2013sinkhorn}, ensuring consistent feature semantics across modalities. Such entropically regularized transport couplings are now widely used as differentiable modules in vision and language model design~\citep{sarlin2020superglue,caron2020swav,izquierdo2024salad,imfeld2024transformer,haviv2024wormhole,xie2025mhc,yang2026mhclite}. To our knowledge, we are the first to adopt this transport-based band alignment for semantic correspondence across heterogeneous remote-sensing spectral bands within a unified low-level restoration foundation model. Second, three complementary mixture-of-experts layers process spatial patterns, spectral fidelity, and global context, enabling specialized yet coordinated feature transformations. Third, step-level dynamic weight adjustment stabilizes joint training across tasks with heterogeneous loss scales and convergence rates.

We conduct comprehensive experiments comparing LLaRS against strong image restoration baselines under a unified training protocol, evaluating both pixel-level metrics and spectral fidelity metrics. Ablation studies quantify the contribution of each architectural component. Finally, we investigate parameter-efficient finetuning of LLaRS on held-out downstream benchmarks, demonstrating that adapting a small fraction of parameters achieves competitive performance with full finetuning while drastically reducing computational cost.

\section{Related Work}
\label{sec:related}

\paragraph{Remote Sensing Foundation Models}
Recent advancements in remote sensing foundation models have been primarily driven by the proliferation of large-scale datasets and the evolution of network architectures. With the continuous expansion of Earth observation data, researchers have constructed massive benchmark datasets encompassing diverse sensors, spatial scales, and modalities. Datasets such as SatLas~\citep{SatLas} and SkySense~\citep{SkySense} provide extensive multi-source imagery, while vision-language datasets like RS5M~\citep{RS5M} and ChatEarthNet~\citep{ChatEarthNet} introduce textual descriptions to facilitate cross-modal learning. Architecturally, there is a distinct paradigm shift from convolutional neural networks to vision Transformers (ViTs)~\citep{Transformer, ViT}. ViTs have become the dominant backbone for remote sensing foundation models due to their exceptional capability in modeling long-range spatial correlations and global contextual dependencies.

In terms of training paradigms, Masked Image Modeling (MIM) and Contrastive Learning (CLR) have emerged as the standard strategies for self-supervised pre-training. Models including SatMAE~\citep{SatMAE}, TerraFM~\citep{TerraFM}, and RingMo~\citep{RingMo} utilize MIM to extract universal visual representations from unannotated imagery by predicting masked patches. While these foundation models have achieved remarkable success in high-level semantic tasks such as scene classification and semantic segmentation, their pre-training objectives are fundamentally designed to capture global semantics. MIM inherently discards substantial pixel information during the masking process, which directly conflicts with the dense reconstruction requirements of low-level vision tasks. Consequently, the applicability and effectiveness of current remote sensing foundation models for pixel-level image restoration remain unverified and present a significant research gap.

\paragraph{Low-level Vision in Remote Sensing}
Low-level vision in remote sensing focuses on pixel-level image restoration and enhancement, encompassing critical processes such as super-resolution, denoising, declouding, and spatiotemporal fusion. These operations are essential for mitigating quality degradation caused by atmospheric disturbances and sensor limitations, ultimately delivering high-quality data for subsequent semantic interpretation. Early approaches typically designed task-specific models or utilized multi-branch networks with decoupled encoding and decoding heads to handle different degradations~\citep{DualCNN, Li_2020_CVPR}. Recently, the field has gravitated towards unified multi-task architectures. Methods like AirNet~\citep{AirNet} and TransWeather~\citep{TransWeather} employ single-encoder single-decoder structures to process diverse degradations simultaneously without requiring prior task knowledge. Furthermore, prompt-driven and instruction-guided frameworks, such as PromptIR~\citep{PromptIR} and InstructIR~\citep{InstructIR}, have demonstrated remarkable cross-scenario adaptability by dynamically modulating network features based on implicit condition vectors or natural language instructions. Other works like DA-CLIP~\citep{DA-CLIP1} leverage vision-language pre-training to achieve cross-modal alignment for generalized image restoration.

Despite these innovations, pre-training foundation models specifically for remote sensing low-level vision faces severe bottlenecks. Current low-level pre-training frameworks, such as IPT~\citep{IPT} and DegAE~\citep{DegAE}, heavily rely on synthesizing degradations on natural image datasets like ImageNet~\citep{ImageNet}. These methods fail to capture the multi-spectral, multi-scale, and multi-sensor complexities inherent to remote sensing imagery. Additionally, there is a critical scarcity of large-scale multi-task sample libraries dedicated to remote sensing image restoration, forcing models to overfit to limited in-domain datasets. When handling multi-source data, existing models often struggle with input misalignment across different spectral channel configurations and spatial sampling intervals, leading to semantic ambiguity. Developing unified architectures that inherently align heterogeneous multi-modal inputs and designing novel training paradigms specifically for pixel-level dense prediction remain unresolved challenges in this domain.
\section{Methodology}
\label{sec:method}

\begin{figure*}[t]
    \centering
    \includegraphics[width=\textwidth]{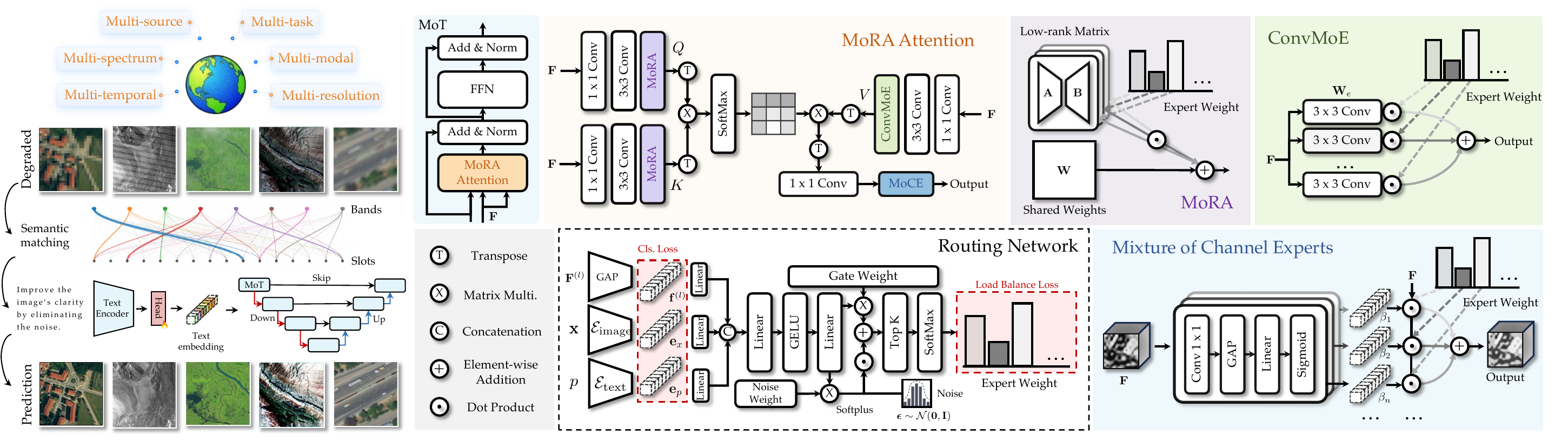}
    \caption{Overall architecture of LLaRS.}
    \label{fig:net-architecture}
\end{figure*}

To address the profound differences in degradation mechanisms, feature hierarchies, and restoration strategies across diverse remote sensing low-level vision tasks, we propose LLaRS (\cref{fig:net-architecture}), a multi-task foundation model based on the mixture-of-experts (MoE)~\citep{MoE} architecture. LLaRS resolves the core contradiction between multi-task commonality and specificity by embedding heterogeneous expert modules into a shared backbone~\citep{UNet}. Given a degraded remote sensing image $\mathbf{x} \in \mathbb{R}^{ C \times  H \times W}$ and a task description text $p$, the model learns a mapping $\mathcal{F}: (\mathbf{x}, p) \mapsto \hat{\mathbf{y}}$, where $\hat{\mathbf{y}} \in \mathbb{R}^{C \times  H \times W}$ represents the restored high-quality image. Here $C$ is a fixed unified channel count: heterogeneous inputs are padded to $C$ before the backbone so batches share a common layout and parameters. The shared backbone captures universal image restoration priors, while the dynamically routed experts learn task-specific strategies without introducing prohibitive computational overhead.

\subsection{Band Semantic Matching}
\label{subsec:sinkhorn}

\begin{figure}[t]
    \centering
    \includegraphics[width=\linewidth]{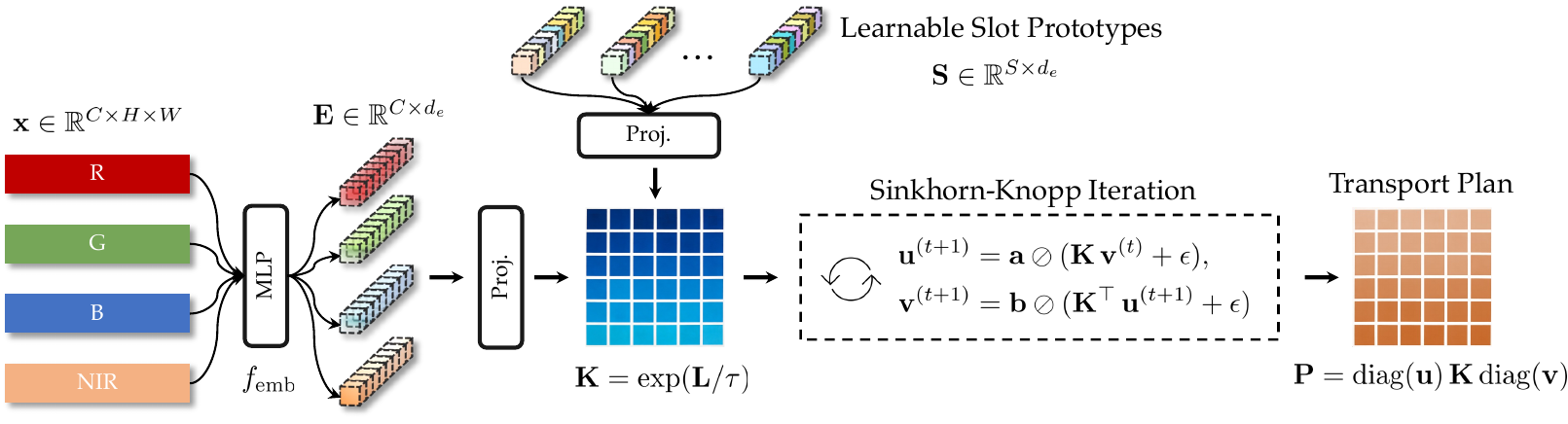}
    \caption{Entropy-regularized channel-to-slot matching.}
    \label{fig:sinkhorn-architecture}
\end{figure}

We propose a band alignment mechanism (\cref{fig:sinkhorn-architecture}) using optimal transport (OT)~\citep{sinkhorn1967concerning,cuturi2013sinkhorn} before the restoration backbone so multi-sensor inputs are routed from their native channel ordering into a fixed bank of $S$ learnable band prototypes. The downstream network thus sees a channel axis with consistent roles instead of raw sensor indices. Across tasks and sensors, both band counts and physical meanings differ; even when all inputs share the same tensor width $C$, the stacks remain semantically heterogeneous. The core difficulty is index semantics: the same band index can correspond to green in one pipeline and a short-wave band in another, so shared early layers cannot treat a band as a stable physical quantity. The same tensor position can also be broadband and texture-rich in one acquisition setting yet a narrow spectral band in another (e.g., pansharpening versus multispectral products), so matching must be driven by per-channel appearance from a shared embedder, not by index alone.

Learnable slot prototypes $\mathbf{S}$ differentiate during training to represent distinct spectral or sensor modes. Turning similarities into a valid coupling requires constraints on both sides: slot-side-only normalization, \ie, including attention that softmaxes over slots, does not impose a channel marginal, so some channels could be oversubscribed by multiple slots while others receive negligible mass. Entropy-regularized optimal transport with Sinkhorn iterations enforces dual marginals so mass is not ignored on either channels or slots and no slot degenerates, which purely slot-wise renormalization does not guarantee~\citep{sinkhorn1967concerning,cuturi2013sinkhorn}. This semantic standardization module is fully differentiable and feeds a shared restoration backbone.

Let $\mathbf{x} \in \mathbb{R}^{C \times H \times W}$ be the channel-first input tensor. A shared lightweight network $f_{\text{emb}}$ maps $\mathbf{x}$ to channel embeddings; slot prototypes $\mathbf{S} \in \mathbb{R}^{S \times d_e}$ and projections $\mathbf{W}_S, \mathbf{W}_E \in \mathbb{R}^{d_e \times d_p}$ define matching scores:
\begin{equation}
\mathbf{E} = f_{\text{emb}}(\mathbf{x}) \in \mathbb{R}^{C \times d_e},
\label{eq:chan-embed}
\end{equation}
\begin{equation}
\mathbf{L} = \frac{(\mathbf{S}\mathbf{W}_S)(\mathbf{E}\mathbf{W}_E)^\top}{\sqrt{d_p}} \in \mathbb{R}^{S \times C}.
\label{eq:logits}
\end{equation}

We formulate an entropic OT problem with uniform marginals $a_s = 1/S$ on slots and $b_c = 1/C$ on channels so mass spreads evenly across both sides. Temperature $\tau$ controls the sharpness of the Gibbs kernel: smaller $\tau$ concentrates mass on high-similarity slot--channel pairs, while larger $\tau$ yields smoother assignments. The entropic kernel $\mathbf{K} \in \mathbb{R}^{S \times C}$ is
\begin{equation}
\mathbf{K} = \exp(\mathbf{L} / \tau).
\label{eq:gibbs-kernel}
\end{equation}
We solve it by iterative Sinkhorn scaling~\citep{sinkhorn1967concerning,cuturi2013sinkhorn}, \ie, alternating row and column normalizations of $\mathbf{K}$:
\begin{equation}
\begin{aligned}
\mathbf{u}^{(t+1)} &= \mathbf{a} \oslash (\mathbf{K}\,\mathbf{v}^{(t)} + \epsilon), \\
\mathbf{v}^{(t+1)} &= \mathbf{b} \oslash (\mathbf{K}^\top\mathbf{u}^{(t+1)} + \epsilon).
\end{aligned}
\label{eq:sinkhorn-iter}
\end{equation}
Here $\oslash$ is element-wise division; we set $\mathbf{v}^{(0)}=\mathbf{1}_C$. In practice, we run the iteration in the log space to avoid numerical instability.
After convergence, with dual potentials $\mathbf{u} \in \mathbb{R}^{S}$ and $\mathbf{v} \in \mathbb{R}^{C}$, the soft assignment matrix is
\begin{equation}
\mathbf{P} = \operatorname{diag}(\mathbf{u})\,\mathbf{K}\,\operatorname{diag}(\mathbf{v}) \in \mathbb{R}^{S \times C}.
\label{eq:transport-plan}
\end{equation}
This alternating scheme generalizes the classical Sinkhorn--Knopp procedure: in the square uniform-marginal case it converges to a doubly stochastic matrix, whereas here it matches the prescribed slot and channel marginals on $\mathbf{K}$.
Applying $\mathbf{P}$ along the channel mode of the same $\mathbf{x}$ yields slot-aligned maps
\begin{equation}
\mathbf{Z} = \mathbf{P}\,\mathbf{x} \in \mathbb{R}^{S \times H \times W}.
\label{eq:routing}
\end{equation}
This differentiable channel--slot coupling learns a data-dependent mapping and projects heterogeneous inputs into a standardized representation for the downstream restoration network.

\subsection{Task-Conditioned Mixture-of-Experts}
\label{subsec:moe-design}

The quality of the routing mechanism directly determines the model's ability to activate the most appropriate expert combination for specific degradations. We design a task-conditioned dynamic routing generator that aggregates multisource information. The text prompt $p$ and degraded image $\mathbf{x}$ are encoded as
\begin{equation}
\mathbf{e}_p = \mathcal{E}_{\text{text}}(p) \in \mathbb{R}^{d_{\text{text}}}, \qquad
\mathbf{e}_x = \mathcal{E}_{\text{image}}(\mathbf{x}) \in \mathbb{R}^{d_{\text{image}}}.
\label{eq:text-image-enc}
\end{equation}
For the current feature map $\mathbf{F}^{(l)} \in \mathbb{R}^{H_l \times W_l \times C_l}$, we apply global average pooling and a linear map to obtain a context vector $\mathbf{f}^{(l)}$, then linearly project $\mathbf{e}_p$, $\mathbf{e}_x$, and $\mathbf{f}^{(l)}$ to a common width $d$, yielding $\mathbf{z}_p$, $\mathbf{z}_x$, and $\mathbf{z}_f$, respectively. We concatenate these route tokens as
\begin{equation}
\mathbf{z} = [\mathbf{z}_p\,;\,\mathbf{z}_x\,;\,\mathbf{z}_f] \in \mathbb{R}^{3d},
\label{eq:route-fusion}
\end{equation}
and pass $\mathbf{z}$ through a two-layer MLP with activation $\phi$ to obtain a fused vector $\mathbf{h}$. Top-$k$ sparse gating on followed by softmax yields routing weights $\boldsymbol{\alpha} \in \mathbb{R}^{E}$ over experts $e \in \mathcal{K}$.

The Convolutional Mixture-of-Experts (ConvMoE) provides diverse spatial kernels. By linearity of convolution, weighted kernels merge into a single convolution~\citep{DynamicConv, CondConv}:
\begin{equation}
\sum_{e \in \mathcal{K}} \alpha_e (\mathbf{F} \ast \mathbf{W}_e)
= \mathbf{F} \ast \Bigl(\sum_{e \in \mathcal{K}} \alpha_e \mathbf{W}_e\Bigr).
\label{eq:convmoe-fusion}
\end{equation}

Expert biases combine in the same way as $\sum_{e \in \mathcal{K}} \alpha_e \mathbf{b}_e$.

Similarly, the Mixture-of-Channel-Experts (MoCE) applies expert channel attention weights $\boldsymbol{\beta}_e$ and merges them with the same $\boldsymbol{\alpha}$:
\begin{equation}
\sum_{e \in \mathcal{K}} \alpha_e (\boldsymbol{\beta}_e \odot \mathbf{F}) = \Bigl(\sum_{e \in \mathcal{K}} \alpha_e \boldsymbol{\beta}_e\Bigr) \odot \mathbf{F}.
\label{eq:moce-fusion}
\end{equation}

For global dependency modeling, we use Mixture-of-LoRA-Attention (MoRA), which is placed in the Mixture-of-Transformer (MoT) block. Standard multi-expert attention repeats the $O(N_{}^2 d)$ attention cost for every expert, where $N_{}$ is the number of spatial positions, resulting in $O(2 |\mathcal{K}|N_{}^2 d )$ total complexity. MoRA reduces this to $O(2 N_{}d^2 + |\mathcal{K}|N_{}Cr)$ by aggregating expert-specific value projections and LoRA query/key adapters with the routing weights $\boldsymbol{\alpha}$ into tensors $\mathbf{V}$, $\Delta\mathbf{Q}$, and $\Delta\mathbf{K}$, where $r \ll d$ is the low-rank dimension. We use base maps $\mathbf{Q}_0 = \mathbf{F}\mathbf{W}_Q$ and $\mathbf{K}_0 = \mathbf{F}\mathbf{W}_K$. A single attention pass yields
\begin{equation}
\mathbf{Y}_{\text{MoT}} = \mathrm{Softmax}\left(\frac{(\mathbf{Q}_0 + \Delta \mathbf{Q})(\mathbf{K}_0 + \Delta \mathbf{K})^{\top}}{\sqrt{d_k}}\right) \mathbf{V}.
\label{eq:mot}
\end{equation}
The supplementary material also gives the exact per-expert mixture, sums over $e \in \mathcal{K}$, a first-order softmax expansion, and remainder terms.

\subsection{Multi-Task Optimization Strategy}
\label{subsec:mtl}

Remote sensing restoration tasks exhibit severe imbalance in difficulty and convergence speeds, leading to gradient dominance and inconsistent learning paces during joint optimization. Traditional gradient manipulation methods, \eg, PCGrad, CAGrad, and NashMTL~\citep{PCGrad,CAGrad,NashMTL}, calculate independent gradients for each task, which incurs prohibitive computational and memory costs for large models handling numerous tasks simultaneously. To overcome this, we propose a step-level Dynamic Weight Adjustment (DWA) strategy that operates entirely within the forward pass without generating additional gradient graphs.

For $M$ task categories present in a minibatch at step $t$, we track the exponential moving average of the loss $\ell_m^{(t)}$ for each task $m$ using a decay coefficient $\gamma$:
\begin{equation}
\bar{\ell}_m^{(t)} = \gamma \cdot \bar{\ell}_m^{(t-1)} + (1 - \gamma) \cdot \ell_m^{(t)}.
\label{eq:dwa-ema}
\end{equation}
The descent rate compares the current loss to its EMA:
\begin{equation}
r_m^{(t)} = \ell_m^{(t)} / {\bar{\ell}_m^{(t)}}.
\label{eq:dwa-rate}
\end{equation}
A larger $r_m^{(t)}$ suggests stagnation and receives higher weight. Normalized task weights use a temperature-scaled softmax:
\begin{equation}
w_m^{(t)} = M \cdot \frac{\exp(r_m^{(t)} / T_w)}{\sum_{j=1}^{M} \exp(r_j^{(t)} / T_w)}.
\label{eq:dwa-weight}
\end{equation}
For a minibatch of $N$ samples, let $m_i$ denote the task category of sample $i$, and let $\mathbf{y}_i$ denote its aligned ground-truth clean image. The DWA-weighted mean reconstruction loss is
\begin{equation}
\mathcal{L}_{\text{dwa}} = \frac{1}{N} \sum_{i=1}^{N} w_{m_i}^{(t)}\, \mathcal{L}_i.
\label{eq:dwa-loss}
\end{equation}
Auxiliary classification on degradation labels $c$ uses the routing token $\mathbf{z}$:
\begin{equation}
\mathcal{L}_{\text{cls}} = \operatorname{CrossEntropy}\!\left(\mathbf{W}_{\text{cls}}\mathbf{z},\, c\right).
\label{eq:cls-loss}
\end{equation}
To prevent routing collapse we add load balancing $\mathcal{L}_{\text{balance}}$~\citep{MoE}, scaled by the stop-gradient of $\mathcal{L}_{\text{dwa}}$:
\begin{equation}
\mathcal{L}_{\text{total}} = \mathcal{L}_{\text{dwa}} + 0.01 \cdot \operatorname{sg}(\mathcal{L}_{\text{dwa}}) \cdot \mathcal{L}_{\text{balance}} + \mathcal{L}_{\text{cls}},
\label{eq:dwa-total-loss}
\end{equation}
where $\operatorname{sg}(\cdot)$ stops gradients. This balances tasks with negligible extra cost versus equal static weights.

% \begin{table}[htbp]
%     \centering
%     \footnotesize
%     \caption{Comparison of computational overhead for multi-task optimization algorithms.}
%     \resizebox{1\linewidth}{!}{
%     \begin{tabular}{lccc}
%     \toprule
%     Method & Backward passes & Extra params & Gradient memory \\
%     \midrule
%     Equal weight & 1 & 0 & $O(D)$ \\
%     PCGrad~\citep{PCGrad} & $T$ & 0 & $O(TD)$ \\
%     MGDA~\citep{MGDA} & $T$ & 0 & $O(TD)$ \\
%     CAGrad~\citep{CAGrad} & $T$ & 0 & $O(TD)$ \\
%     NashMTL~\citep{NashMTL} & $T$ & 0 & $O(TD)$ \\
%     Step DWA (ours) & 1 & 0 & $O(D)$ \\
%     \bottomrule
%     \end{tabular}
%     }
%     \label{tab:mtl-complexity}
% \end{table}
\section{Experiments}
\label{sec:experiments}

\subsection{Data}
\label{subsec:llars1m-data}

To train the model, we construct LLaRS1M, a million-scale dataset that aggregates public multispectral, panchromatic, and SAR imagery across eleven low-level restoration and enhancement tasks. Samples combine real registered pairs (cloud removal, pansharpening, super-resolution, spatiotemporal fusion, speckle-related denoising) with on-the-fly simulated degradations (denoising, deblurring, destriping, histogram and linear-stretch reversal, brightness adjustment) built from high-quality references. Each training tuple pairs a degraded patch with its target and a natural-language prompt drawn from a large per-task pool; training uses task-balanced minibatch construction. See the supplementary material for more construction details.

\begin{figure}[t]
    \centering
    \includegraphics[width=\linewidth]{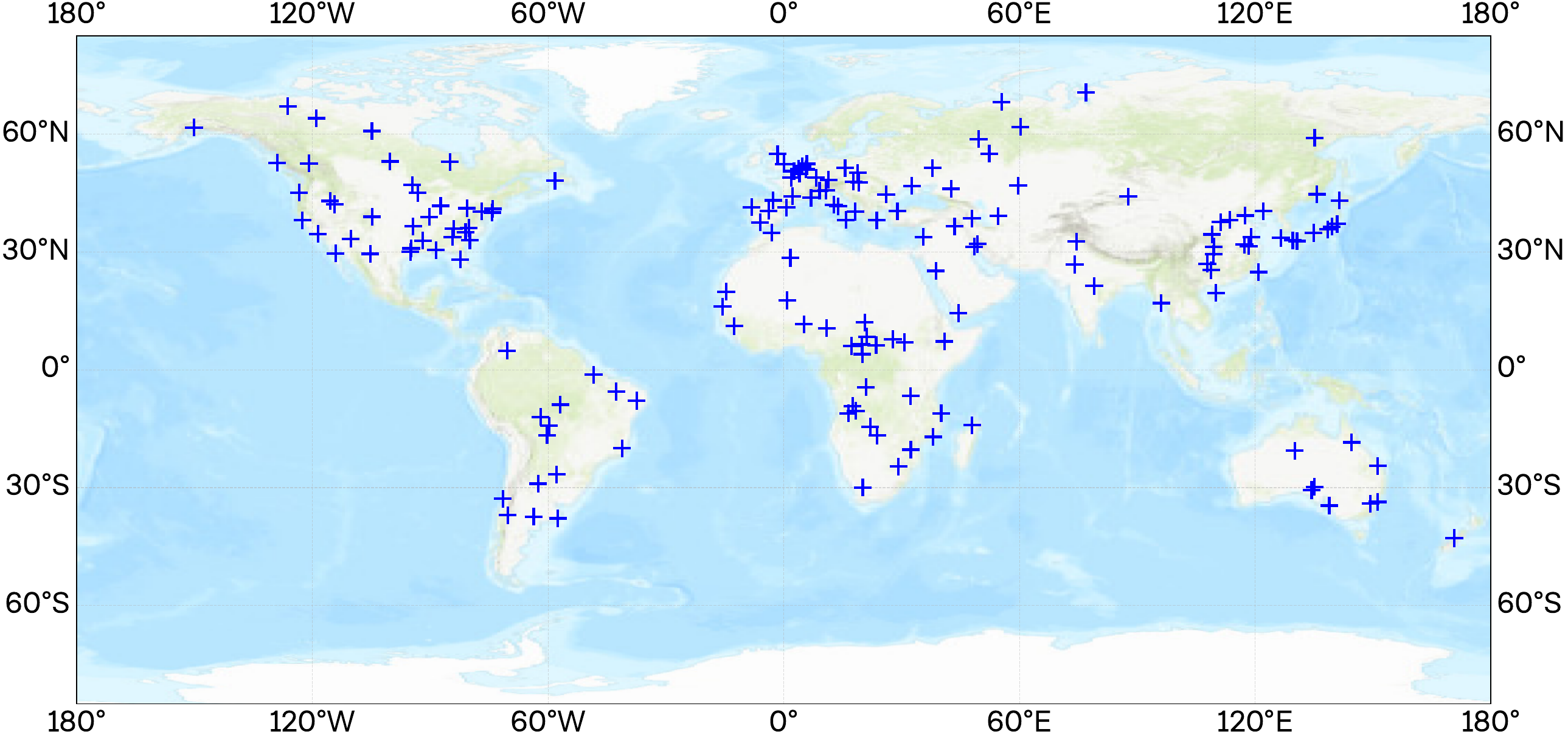}
    \caption{Geographic distribution of LLaRS1M sampling locations.}
    \label{fig:llars1m-geo}
\end{figure}

\cref{fig:llars1m-geo} shows georeferenced tiles spanning continents and major climate belts; wide-area cloud-removal corpora drive much of the footprint. Subsets without geographic metadata (including some synthetic haze imagery and certain downscaled products) do not appear as points, so coverage in practice is broader than the map. Clean patches used to synthesize degradations are drawn from the same paired multisensor pool, inheriting diverse land cover and imaging geometry.

\begin{figure}[t]
    \centering
    \includegraphics[width=\linewidth]{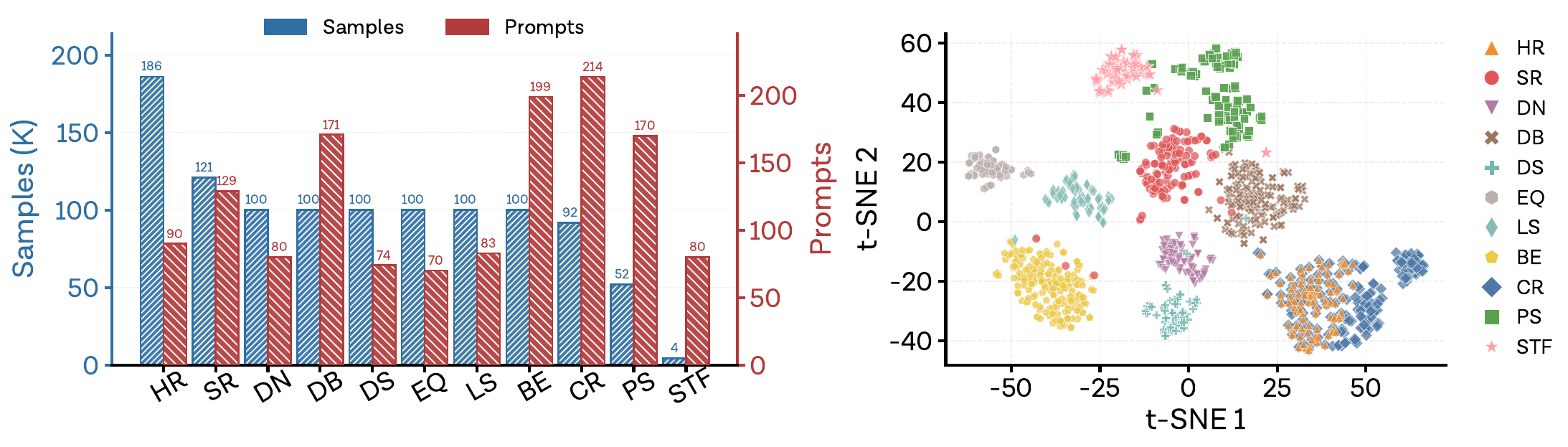}
    \caption{(\textit{Left}) Per-task sample counts and number of prompts per task in LLaRS1M. (\textit{Right}) T-SNE of embedded prompts.}
    \label{fig:llars1m-task-prompt-tsne}
\end{figure}

\cref{fig:llars1m-task-prompt-tsne} (\emph{left}) lists per-task sample totals and prompt counts. The magnitudes follow how each corpus is built: large cropped synthetic dehazing sets and multi-site super-resolution archives contribute high counts, whereas paired Landsat--MODIS series for spatiotemporal fusion are comparatively few; six simulation pipelines each draw a fixed budget from shared clean references. Prompt pool sizes track how richly a task can be verbalized, for instance, cloud removal spans thin versus thick clouds and auxiliary cues, whereas dehaze wording stays closer to a shared lexical core. (\emph{Right}) We encode every prompt and project them with t-SNE~\citep{maaten2008visualizing}. Within-task paraphrases collapse to compact clusters, indicating semantic consistency. Cloud and dehaze clusters abut but remain separable; super-resolution neighbors pansharpening through shared \textit{resolution} language; spatiotemporal fusion drifts toward temporal and fusion vocabulary; embeddings for global radiometry and contrast-style tasks separate from appearance-corruption tasks, \ie, patterns that support language-conditioned routing without collapsing tasks in embedding space.

\begin{figure}[t]
    \centering
    \includegraphics[width=\linewidth]{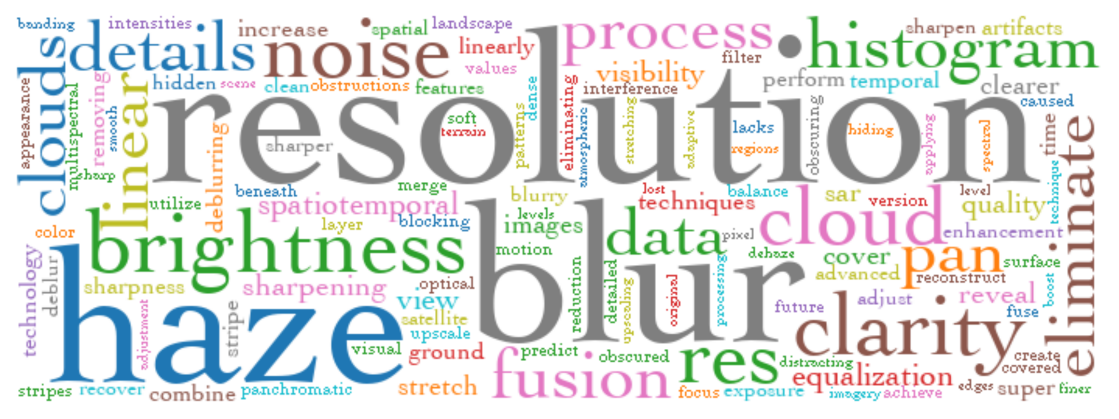}
    \caption{Word cloud of all prompts in LLaRS1M.}
    \label{fig:llars1m-wordcloud}
\end{figure}

\cref{fig:llars1m-wordcloud} highlights tiered vocabulary: high-frequency tokens anchor core degradations (\eg, \textit{haze}, \textit{blur}, \textit{cloud}, \textit{noise}, \textit{brightness}, \textit{clarity}); mid-frequency terms name operations such as \textit{sharpening}, \textit{fusion}, \textit{stripes}, and \textit{histogram}-related wording; rarer words add scene context and intensity nuance. Prompts mix imperatives, symptom-then-request phrasing, questions, and application-grounded descriptions, which broadens instruction styles seen at inference.

% Figure moved to \cref{sec:supp-additional-results}
See \cref{sec:supp-additional-results} for representative LLaRS1M tuples showing visible diversity across all eleven tasks. Training and evaluation details are also provided in \cref{sec:supp-additional-results}.

\subsection{Main Results}

\begin{table}[t]
    \caption{Experimental results: pixel fidelity and structural similarity. \textcolor[HTML]{D6ECF6}{\rule{6pt}{6pt}} Best result. \textcolor[HTML]{F3D7B6}{\rule{6pt}{6pt}} Second-best result.}
    \label{tab:main-psnr-ssim}
    \centering
    \scriptsize
    \setlength{\tabcolsep}{2.5pt}
    \renewcommand{\arraystretch}{1.15}
    \resizebox{1\linewidth}{!}{
    \begin{tabular}{l cc cc cc cc cc cc}
    \toprule
    \multirow{2}{*}{Method} & \multicolumn{2}{c}{CR} & \multicolumn{2}{c}{PS} & \multicolumn{2}{c}{SR} & \multicolumn{2}{c}{HR} & \multicolumn{2}{c}{STF} & \multicolumn{2}{c}{DN} \\
    \cmidrule(lr){2-3}\cmidrule(lr){4-5}\cmidrule(lr){6-7}\cmidrule(lr){8-9}\cmidrule(lr){10-11}\cmidrule(lr){12-13}
    & PSNR & SSIM & PSNR & SSIM & PSNR & SSIM & PSNR & SSIM & PSNR & SSIM & PSNR & SSIM \\
    \midrule
    MPRNet~\citep{MPRNet} & \cellcolor[HTML]{F3D7B6}29.06 & 0.8303 & 32.31 & 0.8928 & 43.89 & 0.9835 & 19.57 & 0.7032 & 32.13 & 0.8962 & 42.38 & 0.9609 \\
    Restormer~\citep{Restormer} & 27.24 & 0.7601 & 34.32 & 0.9346 & 43.96 & 0.9840 & \cellcolor[HTML]{F3D7B6}22.58 & 0.7269 & \cellcolor[HTML]{F3D7B6}32.52 & 0.8969 & 43.66 & 0.9688 \\
    GridFormer~\citep{GridFormer} & 26.60 & 0.6762 & 33.26 & 0.9207 & 43.63 & 0.9831 & 22.07 & 0.7220 & 32.49 & \cellcolor[HTML]{F3D7B6}0.8973 & 43.05 & 0.9656 \\
    PromptIR~\citep{PromptIR} & 26.88 & 0.7636 & \cellcolor[HTML]{F3D7B6}34.46 & \cellcolor[HTML]{F3D7B6}0.9367 & 43.91 & 0.9840 & 22.34 & 0.7218 & 32.21 & \cellcolor[HTML]{F3D7B6}0.8973 & \cellcolor[HTML]{F3D7B6}43.82 & \cellcolor[HTML]{F3D7B6}0.9695 \\
    AMIR~\citep{AMIR} & 28.63 & 0.7628 & 32.90 & 0.9119 & 43.60 & 0.9822 & 21.56 & 0.7154 & 31.91 & 0.8945 & 43.28 & 0.9665 \\
    HOGformer~\citep{HOGformer} & 28.95 & \cellcolor[HTML]{F3D7B6}0.8407 & 32.93 & 0.9086 & \cellcolor[HTML]{F3D7B6}44.02 & \cellcolor[HTML]{F3D7B6}0.9841 & 21.07 & 0.7167 & 32.43 & \cellcolor[HTML]{D6ECF6}0.8987 & 43.22 & 0.9662 \\
    MoCE-IR~\citep{MoCEIR} & 26.98 & 0.7566 & 33.38 & 0.9200 & 43.70 & 0.9836 & 22.16 & \cellcolor[HTML]{F3D7B6}0.7270 & 32.44 & 0.8933 & 42.57 & 0.9633 \\
    LLaRS & \cellcolor[HTML]{D6ECF6}31.18 & \cellcolor[HTML]{D6ECF6}0.8766 & \cellcolor[HTML]{D6ECF6}34.76 & \cellcolor[HTML]{D6ECF6}0.9396 & \cellcolor[HTML]{D6ECF6}44.10 & \cellcolor[HTML]{D6ECF6}0.9844 & \cellcolor[HTML]{D6ECF6}22.81 & \cellcolor[HTML]{D6ECF6}0.7298 & \cellcolor[HTML]{D6ECF6}32.73 & 0.8970 & \cellcolor[HTML]{D6ECF6}43.84 & \cellcolor[HTML]{D6ECF6}0.9699 \\
    \midrule
    \multirow{2}{*}{Method} & \multicolumn{2}{c}{DB} & \multicolumn{2}{c}{DS} & \multicolumn{2}{c}{LS} & \multicolumn{2}{c}{EQ} & \multicolumn{2}{c}{BE} & \multicolumn{2}{c}{Avg.} \\
    \cmidrule(lr){2-3}\cmidrule(lr){4-5}\cmidrule(lr){6-7}\cmidrule(lr){8-9}\cmidrule(lr){10-11}\cmidrule(lr){12-13}
    & PSNR & SSIM & PSNR & SSIM & PSNR & SSIM & PSNR & SSIM & PSNR & SSIM & PSNR & SSIM \\
    \midrule
    MPRNet~\citep{MPRNet} & 38.25 & 0.9639 & 41.82 & 0.9906 & 18.68 & 0.7927 & 17.78 & 0.7542 & 41.46 & 0.9850 & 32.49 & 0.8867 \\
    Restormer~\citep{Restormer} & 40.94 & 0.9766 & 47.31 & 0.9965 & 24.92 & \cellcolor[HTML]{D6ECF6}0.8923 & 18.30 & 0.7627 & \cellcolor[HTML]{F3D7B6}61.02 & \cellcolor[HTML]{F3D7B6}0.9885 & 36.07 & 0.8989 \\
    GridFormer~\citep{GridFormer} & 43.06 & 0.9844 & 45.29 & 0.9951 & 23.34 & 0.8721 & 18.56 & 0.7688 & 56.24 & \cellcolor[HTML]{F3D7B6}0.9885 & 35.24 & 0.8885 \\
    PromptIR~\citep{PromptIR} & \cellcolor[HTML]{F3D7B6}44.33 & \cellcolor[HTML]{F3D7B6}0.9879 & \cellcolor[HTML]{F3D7B6}49.43 & \cellcolor[HTML]{F3D7B6}0.9974 & 23.96 & \cellcolor[HTML]{F3D7B6}0.8900 & 19.16 & \cellcolor[HTML]{F3D7B6}0.7931 & 58.23 & 0.9881 & \cellcolor[HTML]{F3D7B6}36.25 & \cellcolor[HTML]{F3D7B6}0.9027 \\
    AMIR~\citep{AMIR} & 41.61 & 0.9792 & 45.76 & 0.9948 & 22.51 & 0.8458 & 18.56 & 0.7630 & 45.35 & 0.9837 & 34.15 & 0.8909 \\
    HOGformer~\citep{HOGformer} & 42.56 & 0.9834 & 43.55 & 0.9937 & 20.89 & 0.8519 & 18.23 & 0.7838 & 60.78 & \cellcolor[HTML]{D6ECF6}0.9886 & 35.33 & 0.9015 \\
    MoCE-IR~\citep{MoCEIR} & 41.84 & 0.9802 & 43.83 & 0.9930 & \cellcolor[HTML]{F3D7B6}25.03 & 0.8724 & \cellcolor[HTML]{F3D7B6}19.26 & 0.7850 & 57.15 & \cellcolor[HTML]{F3D7B6}0.9885 & 35.31 & 0.8966 \\
    LLaRS & \cellcolor[HTML]{D6ECF6}45.26 & \cellcolor[HTML]{D6ECF6}0.9896 & \cellcolor[HTML]{D6ECF6}49.62 & \cellcolor[HTML]{D6ECF6}0.9976 & \cellcolor[HTML]{D6ECF6}25.29 & 0.8811 & \cellcolor[HTML]{D6ECF6}19.69 & \cellcolor[HTML]{D6ECF6}0.8352 & \cellcolor[HTML]{D6ECF6}64.27 & \cellcolor[HTML]{F3D7B6}0.9885 & \cellcolor[HTML]{D6ECF6}37.60 & \cellcolor[HTML]{D6ECF6}0.9172 \\
    \bottomrule
    \end{tabular}
    }
\end{table}

\begin{table}[t]
    \caption{Experimental results: spectral and radiometric fidelity.\\ \textcolor[HTML]{D6ECF6}{\rule{6pt}{6pt}} Best result. \textcolor[HTML]{F3D7B6}{\rule{6pt}{6pt}} Second-best result.}
    \label{tab:main-spectral}
    \centering
    \scriptsize
    \setlength{\tabcolsep}{2.5pt}
    \renewcommand{\arraystretch}{1.15}
    \resizebox{1\linewidth}{!}{
    \begin{tabular}{l cc cc cc cc cc cc}
    \toprule
    \multirow{2}{*}{Method} & \multicolumn{2}{c}{CR} & \multicolumn{2}{c}{PS} & \multicolumn{2}{c}{SR} & \multicolumn{2}{c}{HR} & \multicolumn{2}{c}{STF} & \multicolumn{2}{c}{DN} \\
    \cmidrule(lr){2-3}\cmidrule(lr){4-5}\cmidrule(lr){6-7}\cmidrule(lr){8-9}\cmidrule(lr){10-11}\cmidrule(lr){12-13}
    & SAM & ERGAS & SAM & ERGAS & SAM & ERGAS & SAM & ERGAS & SAM & ERGAS & SAM & ERGAS \\
    \midrule
    MPRNet~\citep{MPRNet} & 0.1635 & 45.4368 & 0.0902 & 5.9270 & 0.0186 & 1.4666 & 0.0493 & 6.6063 & 0.0747 & 4.4056 & 0.0394 & 1.9584 \\
    Restormer~\citep{Restormer} & 0.1943 & 87.9132 & \cellcolor[HTML]{F3D7B6}0.0742 & 4.5972 & 0.0183 & 1.4408 & \cellcolor[HTML]{F3D7B6}0.0452 & \cellcolor[HTML]{F3D7B6}5.0565 & \cellcolor[HTML]{F3D7B6}0.0719 & 4.2993 & 0.0343 & 1.6787 \\
    GridFormer~\citep{GridFormer} & 0.2519 & 44.1388 & 0.0894 & 5.1833 & 0.0211 & 1.5481 & 0.0458 & 5.3097 & 0.0725 & \cellcolor[HTML]{F3D7B6}4.2417 & 0.0362 & 1.8049 \\
    PromptIR~\citep{PromptIR} & 0.2000 & 26.2611 & 0.0754 & \cellcolor[HTML]{F3D7B6}4.4878 & 0.0192 & 1.5149 & 0.0457 & 5.1578 & \cellcolor[HTML]{D6ECF6}0.0705 & 4.3680 & \cellcolor[HTML]{F3D7B6}0.0336 & \cellcolor[HTML]{D6ECF6}1.6458 \\
    AMIR~\citep{AMIR} & 0.1833 & 98.1772 & 0.0937 & 5.3990 & 0.0198 & 1.5169 & 0.0498 & 5.4899 & 0.0752 & 4.4343 & 0.0367 & 1.7509 \\
    HOGformer~\citep{HOGformer} & \cellcolor[HTML]{F3D7B6}0.1627 & \cellcolor[HTML]{F3D7B6}18.9648 & 0.0837 & 5.4350 & \cellcolor[HTML]{F3D7B6}0.0179 & \cellcolor[HTML]{F3D7B6}1.4333 & 0.0454 & 5.7576 & 0.0728 & 4.3117 & 0.0358 & 1.7682 \\
    MoCE-IR~\citep{MoCEIR} & 0.2173 & 36.0301 & 0.0884 & 5.0966 & 0.0186 & 1.4927 & 0.0453 & 5.3006 & 0.0745 & 4.3262 & 0.0408 & 1.9069 \\
    LLaRS & \cellcolor[HTML]{D6ECF6}0.1215 & \cellcolor[HTML]{D6ECF6}18.9049 & \cellcolor[HTML]{D6ECF6}0.0721 & \cellcolor[HTML]{D6ECF6}4.3447 & \cellcolor[HTML]{D6ECF6}0.0176 & \cellcolor[HTML]{D6ECF6}1.4174 & \cellcolor[HTML]{D6ECF6}0.0442 & \cellcolor[HTML]{D6ECF6}4.9752 & 0.0731 & \cellcolor[HTML]{D6ECF6}4.1268 & \cellcolor[HTML]{D6ECF6}0.0323 & \cellcolor[HTML]{F3D7B6}1.6478 \\
    \midrule
    \multirow{2}{*}{Method} & \multicolumn{2}{c}{DB} & \multicolumn{2}{c}{DS} & \multicolumn{2}{c}{LS} & \multicolumn{2}{c}{EQ} & \multicolumn{2}{c}{BE} & \multicolumn{2}{c}{Avg.} \\
    \cmidrule(lr){2-3}\cmidrule(lr){4-5}\cmidrule(lr){6-7}\cmidrule(lr){8-9}\cmidrule(lr){10-11}\cmidrule(lr){12-13}
    & SAM & ERGAS & SAM & ERGAS & SAM & ERGAS & SAM & ERGAS & SAM & ERGAS & SAM & ERGAS \\
    \midrule
    MPRNet~\citep{MPRNet} & 0.0454 & 2.7054 & 0.0195 & 1.9176 & 0.1925 & 9.5733 & 0.2162 & 7.5153 & 0.0118 & 1.5525 & 0.0837 & 8.0968 \\
    Restormer~\citep{Restormer} & 0.0403 & 2.0639 & 0.0112 & 1.0421 & \cellcolor[HTML]{F3D7B6}0.1315 & 5.0338 & 0.2052 & 7.2706 & 0.0045 & 0.1571 & 0.0755 & 10.9594 \\
    GridFormer~\citep{GridFormer} & 0.0323 & 1.5723 & 0.0155 & 1.2708 & 0.1405 & 5.6150 & 0.1997 & 6.9678 & 0.0091 & 0.3268 & 0.0831 & 7.0890 \\
    PromptIR~\citep{PromptIR} & \cellcolor[HTML]{F3D7B6}0.0273 & \cellcolor[HTML]{F3D7B6}1.3852 & \cellcolor[HTML]{F3D7B6}0.0107 & \cellcolor[HTML]{D6ECF6}0.8082 & \cellcolor[HTML]{D6ECF6}0.1288 & 5.5081 & \cellcolor[HTML]{F3D7B6}0.1948 & 6.7402 & 0.0056 & 0.2334 & \cellcolor[HTML]{F3D7B6}0.0738 & 5.2828 \\
    AMIR~\citep{AMIR} & 0.0369 & 1.8765 & 0.0178 & 1.2428 & 0.1857 & 6.4062 & 0.2043 & 6.9821 & 0.0188 & 1.0011 & 0.0838 & 12.2070 \\
    HOGformer~\citep{HOGformer} & 0.0306 & 1.5560 & 0.0151 & 1.6000 & 0.1585 & 7.3069 & 0.2030 & 7.2328 & \cellcolor[HTML]{D6ECF6}0.0013 & \cellcolor[HTML]{F3D7B6}0.1533 & 0.0752 & \cellcolor[HTML]{F3D7B6}5.0472 \\
    MoCE-IR~\citep{MoCEIR} & 0.0377 & 1.8041 & 0.0246 & 1.5262 & 0.1395 & \cellcolor[HTML]{F3D7B6}5.0269 & 0.1971 & \cellcolor[HTML]{F3D7B6}6.5127 & 0.0075 & 0.2810 & 0.0810 & 6.3004 \\
    LLaRS & \cellcolor[HTML]{D6ECF6}0.0246 & \cellcolor[HTML]{D6ECF6}1.1425 & \cellcolor[HTML]{D6ECF6}0.0081 & \cellcolor[HTML]{F3D7B6}0.8089 & 0.1369 & \cellcolor[HTML]{D6ECF6}4.6598 & \cellcolor[HTML]{D6ECF6}0.1745 & \cellcolor[HTML]{D6ECF6}6.3356 & \cellcolor[HTML]{F3D7B6}0.0032 & \cellcolor[HTML]{D6ECF6}0.1120 & \cellcolor[HTML]{D6ECF6}0.0644 & \cellcolor[HTML]{D6ECF6}4.4069 \\
    \bottomrule
    \end{tabular}
    }
\end{table}

\begin{figure}[t]
    \centering
    \includegraphics[width=\linewidth]{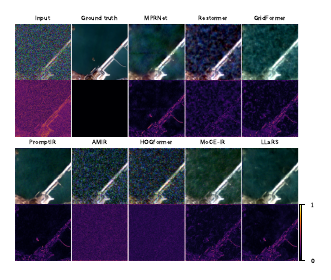}
    \caption{Model predictions and error maps for denoising.}
    \label{fig:qualitative-denoise}
\end{figure}

\begin{figure}[t]
    \centering
    \includegraphics[width=\linewidth]{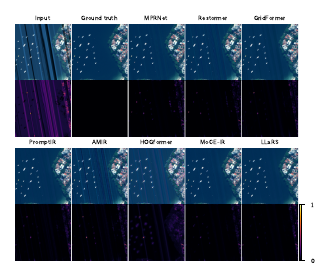}
    \caption{Model predictions and error maps for destriping.}
    \label{fig:qualitative-destripe}
\end{figure}

\begin{figure}[htbp]
    \centering
    \includegraphics[width=\linewidth]{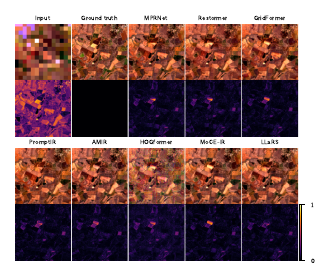}
    \caption{Model predictions and error maps for spatiotemporal fusion.}
    \label{fig:qualitative-stf}
\end{figure}

\cref{tab:main-psnr-ssim,tab:main-spectral} report quantitative results across all eleven restoration tasks. Our approach achieves the highest average performance on PSNR, SSIM, SAM, and ERGAS, demonstrating that the proposed mixture-of-experts architecture with language-conditioned routing effectively handles diverse degradation types within a unified framework. The gains vary by task category, reflecting how different degradation physics interact with the model's architectural components. \cref{fig:qualitative-denoise,fig:qualitative-destripe,fig:qualitative-stf} show qualitative results on denoising, destriping, and spatiotemporal fusion. See \cref{sec:supp-qualitative-results} for additional results.

Expert routing helps most on structured spatial tasks: brightness and destriping see the largest gains (global radiometry vs.\ stripe suppression), while super-resolution, denoising, and deblurring improve by separating frequency- and texture-oriented pathways. Denoising retains more boundary detail than MPRNet~\citep{MPRNet} and PromptIR~\citep{PromptIR}, which over-smooths. On real paired settings, including cloud removal, pansharpening, and spatiotemporal fusion, text prompts and statistical conditioning disentangle heterogeneous sensors, and channel-mixing experts improve spectral fidelity.

Targets such as histogram equalization reversal, dehazing, and linear stretch reversal remain difficult for all methods, as saturation, clipping, and range compression discard information and cap reconstruction quality; accordingly, performance gaps narrow even though our method remains best overall. Destriping benefits from stripe-specific suppression without erasing orthogonal structure; for spatiotemporal fusion, we reduce ghosting where AMIR~\citep{AMIR} and MoCE-IR~\citep{MoCEIR} average temporally inconsistent pixels.

\begin{figure}[t]
    \centering
    \includegraphics[width=\linewidth]{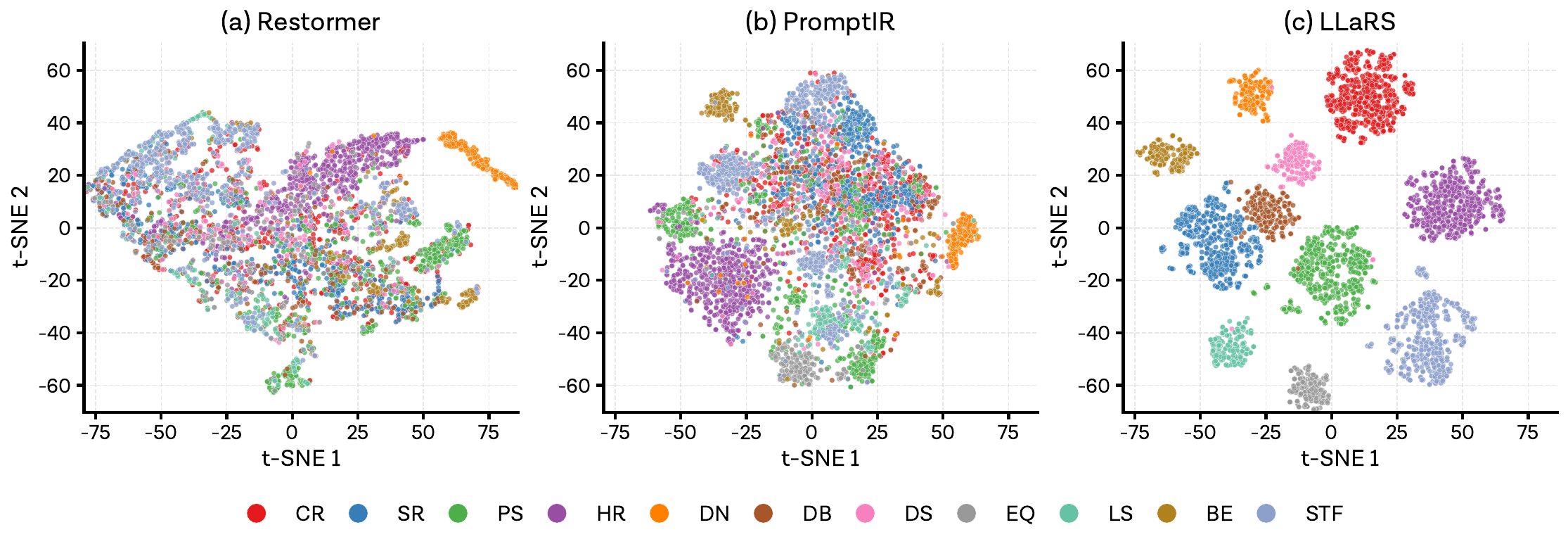}
    \caption{Comparison of t-SNE task feature separability across different models.}
    \label{fig:tsne-comparison}
\end{figure}

Beyond pixel-level reconstruction quality, we examine whether the learned representations exhibit task-specific structure in feature space. Extracting encoder features from the final layer and projecting them via t-SNE reveals distinct clustering patterns across methods (See \cref{fig:tsne-comparison}). Restormer~\citep{Restormer} produces heavily mixed embeddings where different tasks occupy overlapping regions, indicating that the shared backbone learns generic features without strong task specialization. PromptIR~\citep{PromptIR} achieves partial separation, successfully distinguishing dehazing from pansharpening but confusing tasks with similar visual characteristics like different noise types or blur kernels. Our method forms compact, well-separated clusters for each task, with minimal overlap even between visually similar degradations. This separation confirms that the combination of text conditioning and expert routing drives the network toward task-specific feature representations, where different degradation types activate distinct computational pathways through the model. The clear clustering suggests that the gating mechanism has learned meaningful task boundaries rather than arbitrary partitions, providing evidence that the architectural design successfully encourages specialization.

\paragraph{Parameter efficient fine-tuning.}\label{subsec:finetuning}

\begin{figure}[t]
    \centering
    \includegraphics[width=\linewidth]{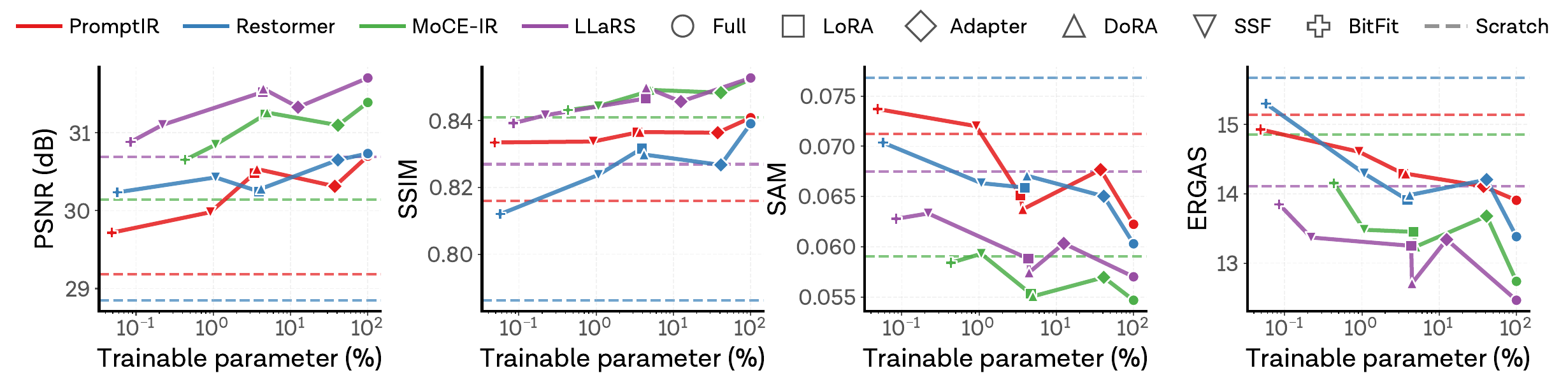}
    \caption{Relationship between trainable parameter ratio and average task performance for different fine-tuning methods.}
    \label{fig:param-metric-trend}
\end{figure}

For PromptIR~\citep{PromptIR}, Restormer~\citep{Restormer}, MoCE-IR~\citep{MoCEIR}, and LLaRS, we compare BitFit~\citep{BitFit}, LoRA~\citep{LoRA}, DoRA~\citep{DoRA}, Adapter~\citep{Adapter}, SSF~\citep{SSF}, and full finetuning. \cref{fig:param-metric-trend} plots metrics versus trainable-parameter ratio on a log scale with dashed from-scratch baselines per model. All curves rise with more trainable parameters and sit above their scratch lines, showing pretraining matters. Slopes are steepest from BitFit to LoRA/DoRA and flatten toward full finetuning, indicating diminishing returns after a few percent of weights.

\subsection{Ablations}
\label{subsec:ablation}

We conduct a series of ablation studies to validate the contribution of each architectural component and training strategy.

\begin{table}[t]
    \caption{Ablation study on mixture-of-experts modules.}
    \label{tab:ablation-moe}
    \centering
    \footnotesize
    \renewcommand{\arraystretch}{1.15}
    \resizebox{1\linewidth}{!}{
    \begin{tabular}{lcccccc}
    \toprule
    Config & PSNR$\uparrow$ & SSIM$\uparrow$ & SAM$\downarrow$ & ERGAS$\downarrow$ & Params(M) & FLOPs(G) \\
    \midrule
    Baseline & 32.76 & 0.8371 & 0.1399 & 14.6733 & 11.85 & 96.1 \\
    + ConvMoE & 34.92 & 0.8900 & 0.0882 & 9.4069 & 75.43 & 96.3 \\
    + ConvMoE + MoCE & 35.08 & 0.8942 & 0.0770 & 5.7072 & 77.15 & 96.8 \\
    + ConvMoE + MoCE + MoT & \textbf{37.60} & \textbf{0.9172} & \textbf{0.0644} & \textbf{4.4069} & 83.64 & 109.6 \\
    \bottomrule
    \end{tabular}
    }
\end{table}

\paragraph{Expert types.}
To understand how different expert types contribute to the overall system, we progressively add ConvMoE, MoCE, and MoT modules on top of a plain U-Net baseline, as shown in \cref{tab:ablation-moe}. The spatial convolution experts in ConvMoE deliver the most substantial initial improvement, raising PSNR and reducing ERGAS significantly. This confirms that learning task-specific spatial filters is crucial for handling diverse degradation patterns, from periodic stripes to localized blur kernels. Adding the channel-mixing experts in MoCE further reduces ERGAS and SAM, particularly benefiting tasks that require preserving spectral correlations across bands. The transformer-based MoT module provides additional gains through global context aggregation, bringing the full model to 37.60 PSNR and 4.41 ERGAS. Importantly, the fused expert implementation prevents FLOPs from scaling linearly with expert count, maintaining computational efficiency even as we increase model capacity through specialization.

\paragraph{OT-based band alignment dynamics.}

\begin{figure*}[t]
    \centering
    \includegraphics[width=\textwidth]{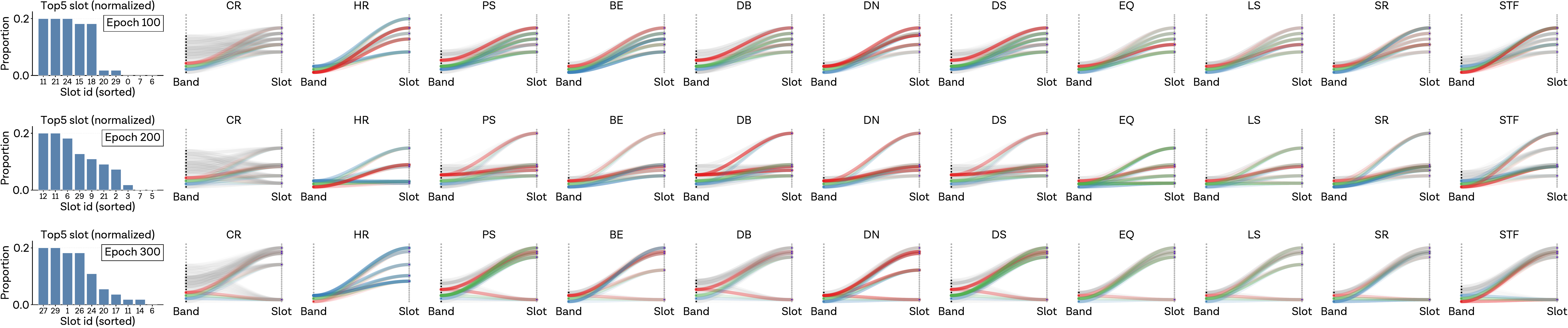}
    \caption{Evolution of channel-to-slot transport for eleven tasks. (\textit{Left}) Top-5 selected slots with the largest channel-to-slot transport mass. (\textit{Right}) The mass flow across channels to slots (the \textcolor{red}{red}, \textcolor{DarkGreen}{green}, and \textcolor{blue}{blue} bands are encoded with colors).}
    \label{fig:sinkhorn-evolution}
\end{figure*}

\cref{fig:sinkhorn-evolution} visualizes how the OT alignment front-end learns stable semantic mappings during training~\citep{sinkhorn1967concerning,cuturi2013sinkhorn}. For clarity, we visualize the top five slots with the largest channel-to-slot transport mass at each checkpoint (ranked by cumulative transported mass). Early in training, couplings remain relatively diffuse, with mass spread across multiple slot candidates per channel. In mid-training, routing preferences sharpen: tasks with similar spectral structure increasingly share slot subsets, while tasks with distinct channel semantics tend to activate disjoint groups. Late in training, assignments stabilize into sharp, task-specific patterns. The utilization histograms (leftmost column), computed on this top-five subset, show that mass does not collapse onto a single slot among them, consistent with the dual-marginal constraint discouraging degenerate routing. Note that several simulation-based tasks may share the same underlying clean references, which can partially couple their semantic content in these visualizations. Tasks with heterogeneous channel counts yield more intricate transport patterns than single-image tasks, yet the learned mappings settle to a consistent structure after convergence. Overall, this progressive specialization suggests that the differentiable entropic OT stage disentangles sensor-specific indexing from semantic content, letting the downstream model operate on a standardized channel representation.

\paragraph{Task weight dynamics.}

\begin{figure}[htbp]
    \centering
    \includegraphics[width=\linewidth]{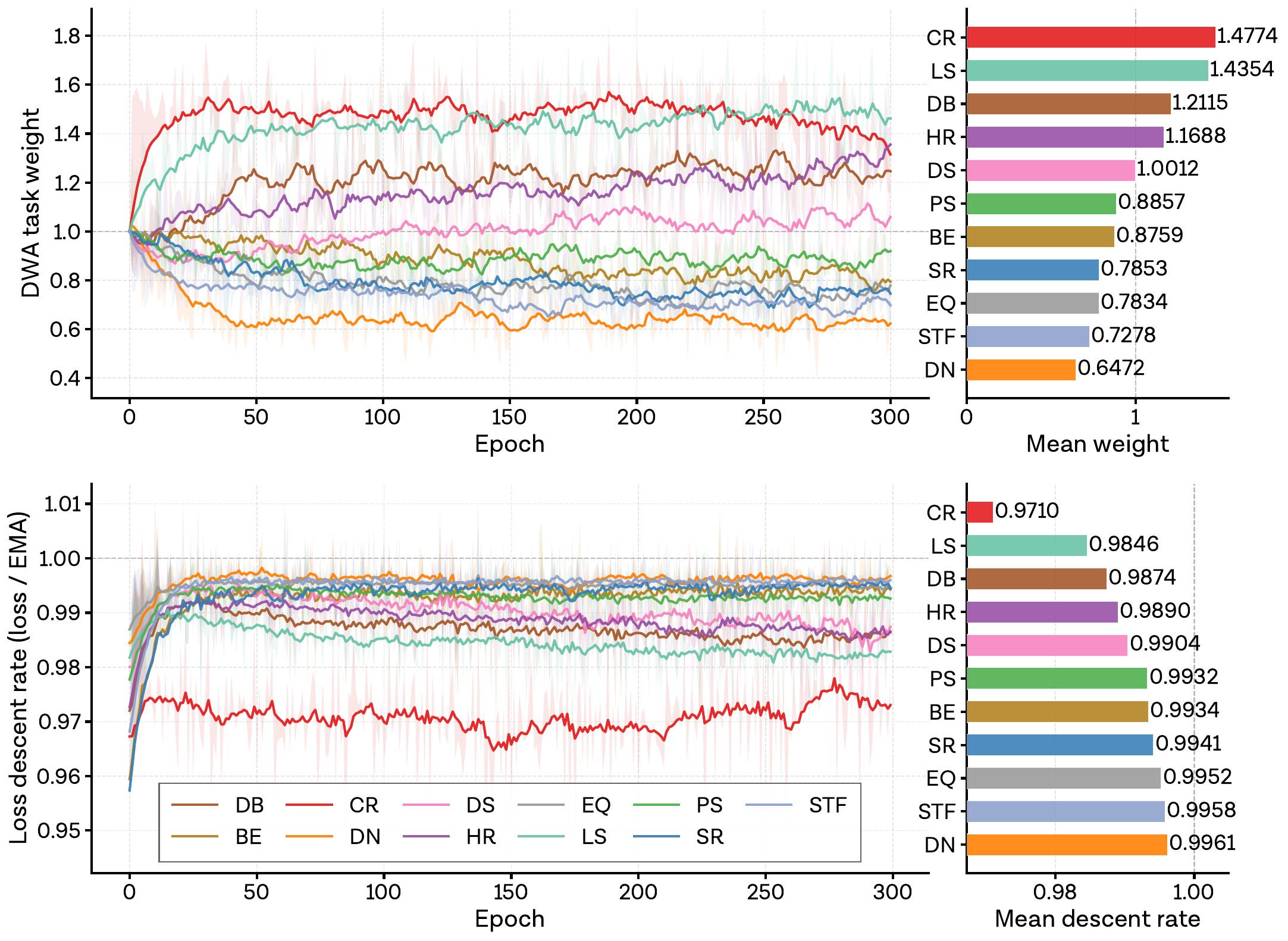}
    \caption{Evolution of per-task weight changing.}
    \label{fig:dwa-rate-evolution}
\end{figure}

\cref{fig:dwa-rate-evolution} reveals the temporal dynamics of task weighting. The top panel shows the evolution of weight changing, while the bottom shows the evolution of loss descent rate. Early epochs exhibit substantial oscillation as the model explores different task combinations. As training progresses, weights stabilize into a clear pattern where challenging tasks like cloud removal and linear stretch reversal maintain higher weights, while tasks that converge more readily such as denoising and brightness adjustment settle to lower weights. This behavior aligns with the design intent of dynamic weight averaging: tasks with slower relative improvement receive higher emphasis to prevent premature convergence on easy tasks at the expense of hard ones.

Additional ablation studies on conditioning mechanisms, the number of experts, multi-task optimization strategies, and model efficiency are provided in \cref{sec:supp-additional-results}.

\section{Conclusion}

This work presents LLaRS, a multi-task foundation model for remote sensing low-level vision. We built LLaRS1M, a large-scale dataset with real pairs and synthetic degradations across eleven restoration tasks, paired with diverse language prompts. Experiments show consistent improvements over strong baselines in pixel and spectral metrics. Parameter-efficient fine-tuning achieves comparable performance to full fine-tuning while the mixture-of-experts architecture degrades gracefully under limited training budgets.

Future work can explore comparing against task-specific small models to better understand the advantages of large multi-task models, and investigating how tasks interact with each other, whether they mutually reinforce or inhibit performance. Additional directions include expanding task coverage toward a unified restoration model, exploring generative approaches for missing information recovery, and joint optimization with downstream semantic tasks to align restoration with application-level objectives.

\clearpage

{\small
\bibliographystyle{ieeenat_fullname}
\bibliography{11_references}

@String(IJCV = {Int. J. Comput. Vis.})

@String(CVPR= {IEEE Conf. Comput. Vis. Pattern Recog.})

@String(ICCV= {Int. Conf. Comput. Vis.})

@String(ECCV= {Eur. Conf. Comput. Vis.})

@String(NIPS= {Adv. Neural Inform. Process. Syst.})

@String(ICLR = {Int. Conf. Learn. Represent.})

@String(AAAI = {AAAI})

@String(IJCV  = {IJCV})

@String(CVPR  = {CVPR})

@String(ICCV  = {ICCV})

@String(ECCV  = {ECCV})

@String(NIPS  = {NeurIPS})

@String(ICLR  = {ICLR})

@String(WACV = {WACV})

@String(MICCAI = {MICCAI})

@String(ESSD = {Earth Syst. Sci. Data})

@String(ACL = {ACL})

@String(SCIS = {Sci. China Inf. Sci.})

@String(JRS = {J. Remote Sens.})

@String(JSTARS = {IEEE J. Sel. Top. Appl. Earth Obs. Remote Sens.})

@String(GRSM = {IEEE Geosci. Remote Sens. Mag.})

@String(TGRS = {IEEE Trans. Geosci. Remote Sens.})

@String(RSE = {Remote Sens. Environ.})

@String(NECO = {Neural Comput.})

@String(PMLR = {Proc. Mach. Learn. Res.})

@String(PJM = {Pac. J. Math.})

@String(JMLR = {J. Mach. Learn. Res.})

@String(DIB = {Data Brief.})

@String(ICML = {ICML})

@article{9082183,
  author   = {Meng, Xiangchao and Xiong, Yiming and Shao, Feng and Shen, Huanfeng and Sun, Weiwei and Yang, Gang and Yuan, Qiangqiang and Fu, Randi and Zhang, Hongyan},
  journal  = GRSM,
  title    = {A Large-Scale Benchmark Data Set for Evaluating Pansharpening Performance: Overview and Implementation},
  year     = {2021},
  volume   = {9},
  number   = {1},
  pages    = {18-52},
  doi      = {10.1109/MGRS.2020.2976696}
}

@inproceedings{Cui_2025_ICCV,
  author    = {Cui, Yongchuan and Liu, Peng and Zhang, Hui},
  title     = {Enpowering Your Pansharpening Models with Generalizability: Unified Distribution is All You Need},
  booktitle = ICCV,
  month     = {October},
  year      = {2025},
  pages     = {11850-11860}
}

@inproceedings{AirNet,
  author    = { Li, Boyun and Liu, Xiao and Hu, Peng and Wu, Zhongqin and Lv, Jiancheng and Peng, Xi },
  booktitle = CVPR,
  title     = {{ All-In-One Image Restoration for Unknown Corruption }},
  year      = {2022},
  pages     = {17431-17441},
  doi       = {10.1109/CVPR52688.2022.01693},
  url       = {https://doi.ieeecomputersociety.org/10.1109/CVPR52688.2022.01693},
  publisher = {IEEE Computer Society},
  address   = {Los Alamitos, CA, USA},
  month     = Jun
}

@inproceedings{AMIR,
  author    = { Yang, Zhiwen and Chen, Haowei and Qian, Ziniu and Yi, Yang and Zhang, Hui and Zhao, Dan and Wei, Bingzheng and Xu, Yan},
  title     = { { All-In-One Medical Image Restoration via Task-Adaptive Routing } },
  booktitle = MICCAI,
  year      = {2024},
  publisher = {Springer Nature Switzerland},
  volume    = {LNCS 15007},
  month     = {October},
  page      = {67 -- 77}
}

@inproceedings{Adapter,
  title     = {Parameter-Efficient Transfer Learning for {NLP}},
  author    = {Houlsby, Neil and Giurgiu, Andrei and Jastrzebski, Stanislaw and Morrone, Bruna and De Laroussilhe, Quentin and Gesmundo, Andrea and Attariyan, Mona and Gelly, Sylvain},
  booktitle = ICML,
  pages     = {2790--2799},
  year      = {2019},
  editor    = {Chaudhuri, Kamalika and Salakhutdinov, Ruslan},
  volume    = {97},
  series    = PMLR,
  month     = {09--15 Jun},
  publisher = {PMLR},
  url       = {https://proceedings.mlr.press/v97/houlsby19a.html}
}

@inproceedings{BitFit,
  title     = {{B}it{F}it: Simple Parameter-efficient Fine-tuning for Transformer-based Masked Language-models},
  author    = {Ben Zaken, Elad  and
               Goldberg, Yoav  and
               Ravfogel, Shauli},
  editor    = {Muresan, Smaranda  and
               Nakov, Preslav  and
               Villavicencio, Aline},
  booktitle = ACL,
  month     = may,
  year      = {2022},
  address   = {Dublin, Ireland},
  publisher = {Association for Computational Linguistics},
  url       = {https://aclanthology.org/2022.acl-short.1/},
  doi       = {10.18653/v1/2022.acl-short.1},
  pages     = {1--9}
}

@inproceedings{CAGrad,
  author    = {Liu, Bo and Liu, Xingchao and Jin, Xiaojie and Stone, Peter and Liu, Qiang},
  title     = {Conflict-averse gradient descent for multi-task learning},
  year      = {2021},
  isbn      = {9781713845393},
  publisher = {Curran Associates Inc.},
  address   = {Red Hook, NY, USA},
  booktitle = NIPS,
  articleno = {1443},
  numpages  = {13},
  series    = {NIPS '21}
}

@article{ChatEarthNet,
  author  = {Yuan, Z. and Xiong, Z. and Mou, L. and Zhu, X. X.},
  title   = {ChatEarthNet: a global-scale image--text dataset empowering vision--language geo-foundation models},
  journal = ESSD,
  volume  = {17},
  year    = {2025},
  number  = {3},
  pages   = {1245--1263},
  url     = {https://essd.copernicus.org/articles/17/1245/2025/},
  doi     = {10.5194/essd-17-1245-2025}
}

@inproceedings{DA-CLIP1,
  title     = {Controlling Vision-Language Models for Multi-Task Image Restoration},
  author    = {Ziwei Luo and Fredrik K. Gustafsson and Zheng Zhao and Jens Sj{\"o}lund and Thomas B. Sch{\"o}n},
  booktitle = ICLR,
  year      = {2024},
  url       = {https://openreview.net/forum?id=t3vnnLeajU}
}

@inproceedings{DegAE,
  author    = {Liu, Yihao and He, Jingwen and Gu, Jinjin and Kong, Xiangtao and Qiao, Yu and Dong, Chao},
  booktitle = CVPR,
  title     = {DegAE: A New Pretraining Paradigm for Low-Level Vision},
  year      = {2023},
  volume    = {},
  number    = {},
  pages     = {23292-23303},
  doi       = {10.1109/CVPR52729.2023.02231}
}

@inproceedings{DualCNN,
  author    = {Pan, Jinshan and Liu, Sifei and Sun, Deqing and Zhang, Jiawei and Liu, Yang and Ren, Jimmy and Li, Zechao and Tang, Jinhui and Lu, Huchuan and Tai, Yu-Wing and Yang, Ming-Hsuan},
  booktitle = CVPR,
  title     = {Learning Dual Convolutional Neural Networks for Low-Level Vision},
  year      = {2018},
  volume    = {},
  number    = {},
  pages     = {3070-3079},
  doi       = {10.1109/CVPR.2018.00324}
}

@inproceedings{DoRA,
  author    = {Liu, Shih-Yang and Wang, Chien-Yi and Yin, Hongxu and Molchanov, Pavlo and Wang, Yu-Chiang Frank and Cheng, Kwang-Ting and Chen, Min-Hung},
  title     = {DoRA: weight-decomposed low-rank adaptation},
  year      = {2024},
  publisher = {JMLR.org},
  booktitle = ICML,
  articleno = {1299},
  numpages  = {22},
  location  = {Vienna, Austria},
  series    = {ICML'24}
}

@article{GridFormer,
  title     = {{GridFormer}: Residual dense transformer with grid structure for image restoration in adverse weather conditions},
  author    = {Wang, Tao and Zhang, Kaihao and Shao, Ziqian and Luo, Wenhan and Stenger, Bjorn and Lu, Tong and Kim, Tae-Kyun and Liu, Wei and Li, Hongdong},
  journal   = IJCV,
  volume    = {132},
  number    = {10},
  pages     = {4541--4563},
  year      = {2024},
  publisher = {Springer}
}

@inproceedings{GradNorm,
  title     = {{G}rad{N}orm: Gradient Normalization for Adaptive Loss Balancing in Deep Multitask Networks},
  author    = {Chen, Zhao and Badrinarayanan, Vijay and Lee, Chen-Yu and Rabinovich, Andrew},
  booktitle = ICML,
  pages     = {794--803},
  year      = {2018},
  editor    = {Dy, Jennifer and Krause, Andreas},
  volume    = {80},
  series    = PMLR,
  month     = {10--15 Jul},
  publisher = {PMLR},
  url       = {https://proceedings.mlr.press/v80/chen18a.html}
}

@article{HOGformer,
  title        = {Gradient as Conditions: Rethinking {HOG} for All-in-one Image Restoration},
  volume       = {40},
  url          = {https://ojs.aaai.org/index.php/AAAI/article/view/38042},
  doi          = {10.1609/aaai.v40i13.38042},
  number       = {13},
  journal      = AAAI,
  author       = {Wu, Jiawei and Yang, Zhifei and Wang, Zhe and Jin, Zhi},
  year         = {2026},
  month        = {Mar.},
  pages        = {10682-10690}
}

@article{TerraFM,
  title         = {{TerraFM}: A Scalable Foundation Model for Unified Multisensor Earth Observation},
  author        = {Muhammad Sohail Danish and Muhammad Akhtar Munir and Syed Roshaan Ali Shah and Muhammad Haris Khan and Rao Muhammad Anwer and Jorma Laaksonen and Fahad Shahbaz Khan and Salman Khan},
  year          = {2025},
  eprint        = {2506.06281},
  archiveprefix = {arXiv},
  journal       = {arXiv},
  primaryclass  = {cs.CV},
  url           = {https://arxiv.org/abs/2506.06281}
}

@INPROCEEDINGS {ImageNet,
author = { Deng, Jia and Dong, Wei and Socher, Richard and Li, Li-Jia and Kai Li and Li Fei-Fei },
booktitle = CVPR,
title = {{ {ImageNet}: A large-scale hierarchical image database }},
year = {2009},
volume = {},
ISSN = {1063-6919},
pages = {248-255},
doi = {10.1109/CVPR.2009.5206848},
url = {https://doi.ieeecomputersociety.org/10.1109/CVPR.2009.5206848},
publisher = {IEEE Computer Society},
address = {Los Alamitos, CA, USA},
month =Jun}

@inproceedings{InstructIR,
author = {Conde, Marcos V. and Geigle, Gregor and Timofte, Radu},
title = {{InstructIR}: High-Quality Image Restoration Following Human Instructions},
year = {2024},
isbn = {978-3-031-72763-4},
publisher = {Springer-Verlag},
address = {Berlin, Heidelberg},
url = {https://doi.org/10.1007/978-3-031-72764-1_1},
doi = {10.1007/978-3-031-72764-1_1},
booktitle = ECCV,
pages = {1–21},
numpages = {21},
location = {Milan, Italy}
}

@INPROCEEDINGS{IPT,
  author={Chen, Hanting and Wang, Yunhe and Guo, Tianyu and Xu, Chang and Deng, Yiping and Liu, Zhenhua and Ma, Siwei and Xu, Chunjing and Xu, Chao and Gao, Wen},
  booktitle=CVPR, 
  title={Pre-Trained Image Processing Transformer}, 
  year={2021},
  volume={},
  number={},
  pages={12294-12305},
  doi={10.1109/CVPR46437.2021.01212}}

@INPROCEEDINGS{Li_2020_CVPR,
  author={Li, Ruoteng and Tan, Robby T. and Cheong, Loong-Fah},
  booktitle=CVPR, 
  title={All in One Bad Weather Removal Using Architectural Search}, 
  year={2020},
  volume={},
  number={},
  pages={3172-3182},
  doi={10.1109/CVPR42600.2020.00324}}

@inproceedings{LoRA,
  title     = {{LoRA}: Low-Rank Adaptation of Large Language Models},
  author    = {Hu, Edward J and Shen, Yelong and Wallis, Phillip and Allen-Zhu, Zeyuan and Li, Yuanzhi and Wang, Shean and Wang, Lu and Chen, Weizhu},
  booktitle = ICLR,
  year      = {2022}
}

@inproceedings{MGDA,
author = {Sener, Ozan and Koltun, Vladlen},
title = {Multi-task learning as multi-objective optimization},
year = {2018},
publisher = {Curran Associates Inc.},
address = {Red Hook, NY, USA},
booktitle = NIPS,
pages = {525–536},
numpages = {12},
location = {Montr\'{e}al, Canada},
series = {NIPS'18}
}

@INPROCEEDINGS{MPRNet,
  author={Mehri, Armin and Ardakani, Parichehr B. and Sappa, Angel D.},
  booktitle=WACV, 
  title={{MPRNet}: Multi-Path Residual Network for Lightweight Image Super Resolution}, 
  year={2021},
  volume={},
  number={},
  pages={2703-2712},
  doi={10.1109/WACV48630.2021.00275}}

@InProceedings{MoCEIR,
    author    = {Zamfir, Eduard and Wu, Zongwei and Mehta, Nancy and Tan, Yuedong and Paudel, Danda Pani and Zhang, Yulun and Timofte, Radu},
    title     = {Complexity Experts are Task-Discriminative Learners for Any Image Restoration},
    booktitle = CVPR,
    month     = {June},
    year      = {2025},
    pages     = {12753-12763}
}

@article{MoE,
  author   = {Jacobs, Robert A. and Jordan, Michael I. and Nowlan, Steven J. and Hinton, Geoffrey E.},
  journal  = NECO,
  title    = {Adaptive Mixtures of Local Experts},
  year     = {1991},
  volume   = {3},
  number   = {1},
  pages    = {79-87},
  doi      = {10.1162/neco.1991.3.1.79}
}

@inproceedings{NashMTL,
  title        = {Multi-task learning as a bargaining game},
  author       = {Navon, Aviv and Shamsian, Aviv and Achituve, Idan and Maron, Haggai and Kawaguchi, Kenji and Chechik, Gal and Fetaya, Ethan},
  booktitle    = ICML,
  pages        = {16428--16446},
  year         = {2022},
  organization = {PMLR}
}

@article{OLI2MSI,
  title     = {Multisensor Remote Sensing Imagery Super-Resolution with Conditional GAN},
  author    = {Wang, Junwei and Gao, Kun and Zhang, Zhenzhou and Ni, Chong and Hu, Zibo and Chen, Dayu and Wu, Qiong},
  journal   = JRS,
  volume    = {2021},
  year      = {2021},
  publisher = {AAAS}
}

@inproceedings{PCGrad,
author = {Yu, Tianhe and Kumar, Saurabh and Gupta, Abhishek and Levine, Sergey and Hausman, Karol and Finn, Chelsea},
title = {Gradient surgery for multi-task learning},
year = {2020},
isbn = {9781713829546},
publisher = {Curran Associates Inc.},
address = {Red Hook, NY, USA},
booktitle = NIPS,
articleno = {489},
numpages = {13},
location = {Vancouver, BC, Canada},
series = {NIPS '20}
}

@inproceedings{PromptIR,
author = {Potlapalli, Vaishnav and Zamir, Syed Waqas and Khan, Salman and Khan, Fahad Shahbaz},
title = {{PromptIR}: prompting for all-in-one blind image restoration},
year = {2023},
publisher = {Curran Associates Inc.},
address = {Red Hook, NY, USA},
booktitle = NIPS,
articleno = {3121},
numpages = {19},
location = {New Orleans, LA, USA},
series = {NIPS '23}
}

@article{RingMo,
  author  = {Sun, Xian and Wang, Peijin and Lu, Wanxuan and Zhu, Zicong and Lu, Xiaonan and He, Qibin and Li, Junxi and Rong, Xuee and Yang, Zhujun and Chang, Hao and He, Qinglin and Yang, Guang and Wang, Ruiping and Lu, Jiwen and Fu, Kun},
  journal = TGRS,
  title   = {{RingMo}: A Remote Sensing Foundation Model With Masked Image Modeling},
  year    = {2023},
  volume  = {61},
  pages   = {1-22}
}

@article{RS5M,
  author  = {Zhang, Zilun and Zhao, Tiancheng and Guo, Yulong and Yin, Jianwei},
  journal = TGRS,
  title   = {{RS5M} and {GeoRSCLIP}: A Large Scale Vision-Language Dataset and A Large Vision-Language Model for Remote Sensing},
  year    = {2024},
  volume  = {62},
  pages   = {1-23}
}

@misc{RICE,
  title         = {A Remote Sensing Image Dataset for Cloud Removal},
  author        = {Daoyu Lin and Guangluan Xu and Xiaoke Wang and Yang Wang and Xian Sun and Kun Fu},
  year          = {2019},
  eprint        = {1901.00600},
  archiveprefix = {arXiv},
  primaryclass  = {cs.CV},
  url           = {https://arxiv.org/abs/1901.00600}
}

@inproceedings{Restormer,
  title     = {Restormer: {E}fficient transformer for high-resolution image restoration},
  author    = {Zamir, Syed Waqas and Arora, Aditya and Khan, Salman and Hayat, Munawar and Khan, Fahad Shahbaz and Yang, Ming-Hsuan},
  booktitle = CVPR,
  pages     = {5728--5739},
  year      = {2022}
}

@article{SARDespeckleDataset,
  author  = {Guan, Jianjun and Zhong, Ping and Zhang, Fan and Liu, Yuhan},
  title   = {Robust {SAR} Image Despeckling by Deep Learning From Near-Real Datasets},
  journal = JSTARS,
  year    = {2024},
  volume  = {17},
  pages   = {3475-3487},
  doi     = {10.1109/JSTARS.2023.3345067}
}

@inproceedings{SatLas,
  author    = {Bastani, Favyen and Wolters, Piper and Gupta, Ritwik and Ferdinando, Joe and Kembhavi, Aniruddha},
  booktitle = ICCV,
  title     = {{SatlasPretrain}: A Large-Scale Dataset for Remote Sensing Image Understanding},
  year      = {2023},
  pages     = {16726-16736}
}

@inproceedings{SatMAE,
author = {Cong, Yezhen and Khanna, Samar and Meng, Chenlin and Liu, Patrick and Rozi, Erik and He, Yutong and Burke, Marshall and Lobell, David B. and Ermon, Stefano},
title = {{SatMAE}: {P}re-training transformers for temporal and multi-spectral satellite imagery},
year = {2022},
isbn = {9781713871088},
publisher = {Curran Associates Inc.},
address = {Red Hook, NY, USA},
booktitle = NIPS,
articleno = {15},
numpages = {15},
location = {New Orleans, LA, USA},
series = {NIPS '22}
}

@inproceedings{SkySense,
  author    = {Guo, Xin and Lao, Jiangwei and Dang, Bo and Zhang, Yingying and Yu, Lei and Ru, Lixiang and Zhong, Liheng and Huang, Ziyuan and Wu, Kang and Hu, Dingxiang and He, Huimei and Wang, Jian and Chen, Jingdong and Yang, Ming and Zhang, Yongjun and Li, Yansheng},
  booktitle = CVPR,
  title     = {{SkySense}: A Multi-Modal Remote Sensing Foundation Model Towards Universal Interpretation for Earth Observation Imagery},
  year      = {2024},
  pages     = {27662-27673}
}

@inproceedings{SSF,
author = {Lian, Dongze and Zhou, Daquan and Feng, Jiashi and Wang, Xinchao},
title = {Scaling \& shifting your features: a new baseline for efficient model tuning},
year = {2022},
isbn = {9781713871088},
publisher = {Curran Associates Inc.},
address = {Red Hook, NY, USA},
booktitle = NIPS,
articleno = {9},
numpages = {15},
location = {New Orleans, LA, USA},
series = {NIPS '22}
}

@inproceedings{Transformer,
  author    = {Vaswani, Ashish and Shazeer, Noam and Parmar, Niki and Uszkoreit, Jakob and Jones, Llion and Gomez, Aidan N. and Kaiser, \L{}ukasz and Polosukhin, Illia},
  title     = {Attention is all you need},
  year      = {2017},
  isbn      = {9781510860964},
  publisher = {Curran Associates Inc.},
  address   = {Red Hook, NY, USA},
  booktitle = NIPS,
  pages     = {6000--6010},
  numpages  = {11},
  location  = {Long Beach, California, USA},
  series    = {NIPS'17}
}

@inproceedings{TransWeather,
  author    = {Valanarasu, Jeya Maria Jose and Yasarla, Rajeev and Patel, Vishal M.},
  title     = {TransWeather: Transformer-Based Restoration of Images Degraded by Adverse Weather Conditions},
  booktitle = CVPR,
  year      = {2022},
  pages     = {2353-2363}
}

@InProceedings{UNet,
author="Ronneberger, Olaf
and Fischer, Philipp
and Brox, Thomas",
editor="Navab, Nassir
and Hornegger, Joachim
and Wells, William M.
and Frangi, Alejandro F.",
title={{U-Net}: Convolutional Networks for Biomedical Image Segmentation},
booktitle=MICCAI,
year="2015",
publisher="Springer International Publishing",
address="Cham",
pages="234--241",
isbn="978-3-319-24574-4"
}

@inproceedings{ResNet,
  author    = {He, Kaiming and Zhang, Xiangyu and Ren, Shaoqing and Sun, Jian},
  title     = {Deep Residual Learning for Image Recognition},
  booktitle = CVPR,
  year      = {2016},
  pages     = {770--778}
}

@misc{DistilBERT,
  author        = {Sanh, Victor and Debut, Lysandre and Chaumond, Julien and Wolf, Thomas},
  title         = {{DistilBERT}, a distilled version of {BERT}: smaller, faster, cheaper and lighter},
  year          = {2019},
  eprint        = {1910.01108},
  archiveprefix = {arXiv},
  primaryclass  = {cs.CL},
  url           = {https://arxiv.org/abs/1910.01108}
}

@inproceedings{ViT,
  title     = {An Image is Worth 16x16 Words: Transformers for Image Recognition at Scale},
  author    = {Dosovitskiy, Alexey and Beyer, Lucas and Kolesnikov, Alexander and Weissenborn, Dirk and Zhai, Xiaohua and Unterthiner, Thomas and  Dehghani, Mostafa and Minderer, Matthias and Heigold, Georg and Gelly, Sylvain and Uszkoreit, Jakob and Houlsby, Neil},
  booktitle = ICLR,
  year      = {2021}
}

@inproceedings{WACV2020_Haze1K,
  author    = {Huang, Binghui and Zhi, Li and Yang, Chao and Sun, Fuchun and Song, Yixu},
  title     = {Single Satellite Optical Imagery Dehazing using SAR Image Prior Based on conditional Generative Adversarial Networks},
  booktitle = WACV,
  year      = {2020},
  pages     = {1806-1813},
  doi       = {10.1109/WACV45572.2020.9093471}
}

@misc{he2024diffusionmodelslowlevelvision,
  title         = {Diffusion Models in Low-Level Vision: A Survey},
  author        = {Chunming He and Yuqi Shen and Chengyu Fang and Fengyang Xiao and Longxiang Tang and Yulun Zhang and Wangmeng Zuo and Zhenhua Guo and Xiu Li},
  year          = {2024},
  eprint        = {2406.11138},
  archiveprefix = {arXiv},
  primaryclass  = {cs.CV},
  url           = {https://arxiv.org/abs/2406.11138}
}

@article{li2020spatio,
  title     = {Spatio-temporal fusion for remote sensing data: An overview and new benchmark},
  author    = {Li, Jun and Li, Yunfei and He, Lin and Chen, Jin and Plaza, Antonio},
  journal   = SCIS,
  volume    = {63},
  number    = {4},
  pages     = {140301},
  year      = {2020},
  publisher = {Springer}
}

@misc{lu2024aifoundationmodelsremote,
  title         = {AI Foundation Models in Remote Sensing: A Survey},
  author        = {Siqi Lu and Junlin Guo and James R Zimmer-Dauphinee and Jordan M Nieusma and Xiao Wang and Parker VanValkenburgh and Steven A Wernke and Yuankai Huo},
  year          = {2024},
  eprint        = {2408.03464},
  archiveprefix = {arXiv},
  primaryclass  = {cs.CV},
  url           = {https://arxiv.org/abs/2408.03464}
}

@misc{zhou2024visionlanguagegeofoundationmodelsurvey,
  title         = {Towards Vision-Language Geo-Foundation Model: A Survey},
  author        = {Yue Zhou and Litong Feng and Yiping Ke and Xue Jiang and Junchi Yan and Xue Yang and Wayne Zhang},
  year          = {2024},
  eprint        = {2406.09385},
  archiveprefix = {arXiv},
  primaryclass  = {cs.CV},
  url           = {https://arxiv.org/abs/2406.09385}
}

@article{sinkhorn1967concerning,
  title   = {Concerning Nonnegative Matrices and Doubly Stochastic Matrices},
  author  = {Sinkhorn, Richard and Knopp, Paul},
  journal = PJM,
  volume  = {21},
  number  = {2},
  pages   = {343--348},
  year    = {1967}
}

@inproceedings{cuturi2013sinkhorn,
  title     = {Sinkhorn Distances: Lightspeed Computation of Optimal Transport},
  author    = {Cuturi, Marco},
  booktitle = NIPS,
  pages     = {2292--2300},
  year      = {2013}
}

@inproceedings{sarlin2020superglue,
  title     = {{SuperGlue}: Learning Feature Matching with Graph Neural Networks},
  author    = {Sarlin, Paul-Edouard and DeTone, Daniel and Malisiewicz, Tomasz and Rabinovich, Andrew},
  booktitle = CVPR,
  pages     = {4938--4947},
  year      = {2020}
}

@inproceedings{caron2020swav,
  title     = {Unsupervised Learning of Visual Features by Contrasting Cluster Assignments},
  author    = {Caron, Mathilde and Misra, Ishan and Mairal, Julien and Goyal, Priya and Bojanowski, Piotr and Joulin, Armand},
  booktitle = NIPS,
  volume    = {33},
  pages     = {9912--9924},
  year      = {2020}
}

@inproceedings{izquierdo2024salad,
  title     = {Optimal Transport Aggregation for Visual Place Recognition},
  author    = {Izquierdo, Sergio and Civera, Javier},
  booktitle = CVPR,
  month     = jun,
  pages     = {17658--17668},
  year      = {2024}
}

@inproceedings{imfeld2024transformer,
  title     = {Transformer Fusion with Optimal Transport},
  author    = {Imfeld, Moritz and Graldi, Jacopo and Giordano, Marco and Hofmann, Thomas and Anagnostidis, Sotiris and Singh, Sidak Pal},
  booktitle = ICLR,
  year      = {2024},
  url       = {https://openreview.net/forum?id=LjeqMvQpen}
}

@inproceedings{haviv2024wormhole,
  title     = {Wasserstein Wormhole: Scalable Optimal Transport Distance with Transformer},
  author    = {Haviv, Doron and Kunes, Russell Zhang and Dougherty, Thomas and Burdziak, Cassandra and Nawy, Tal and Gilbert, Anna and Pe'er, Dana},
  booktitle = ICML,
  series    = PMLR,
  volume    = {235},
  pages     = {17697--17718},
  year      = {2024},
  publisher = {PMLR}
}

@article{xie2025mhc,
  title   = {m{HC}: Manifold-Constrained Hyper-Connections},
  author  = {Xie, Zhenda and Wei, Yixuan and Cao, Huanqi and Zhao, Chenggang and Deng, Chengqi and Li, Jiashi and Dai, Damai and Gao, Huazuo and Chang, Jiang and Yu, Kuai and Zhao, Liang and Zhou, Shangyan and Xu, Zhean and Zhang, Zhengyan and Zeng, Wangding and Hu, Shengding and Wang, Yuqing and Yuan, Jingyang and Wang, Lean and Liang, Wenfeng},
  journal = {arXiv},
  year    = {2025},
  eprint  = {2512.24880},
  archiveprefix = {arXiv}
}

@article{yang2026mhclite,
  title   = {m{HC}-lite: {Y}ou {D}on't {N}eed 20 {Sinkhorn-Knopp} {I}terations},
  author  = {Yang, Yongyi and Gao, Jianyang},
  journal = {arXiv},
  year    = {2026},
  eprint  = {2601.05732},
  archiveprefix = {arXiv}
}

@article{maaten2008visualizing,
  title   = {Visualizing Data using {t-SNE}},
  author  = {Maaten, Laurens van der and Hinton, Geoffrey},
  journal = JMLR,
  volume  = {9},
  number  = {86},
  pages   = {2579--2605},
  year    = {2008}
}

@inproceedings{CondConv,
  title     = {Condconv: Conditionally parameterized convolutions for efficient inference},
  author    = {Yang, Brandon and Bender, Gabriel and Le, Quoc V and Ngiam, Jiquan},
  booktitle = NIPS,
  volume    = {32},
  year      = {2019}
}

@inproceedings{DynamicConv,
  title     = {Dynamic convolution: Attention over convolution kernels},
  author    = {Chen, Yinpeng and Dai, Xiyang and Liu, Mengchen and Chen, Dongdong and Yuan, Lu and Liu, Zicheng},
  booktitle = CVPR,
  pages     = {11030--11039},
  year      = {2020}
}

@article{SEN12MS-CR,
  title     = {{Multisensor Data Fusion for Cloud Removal in Global and All-Season Sentinel-2 Imagery}},
  author    = {Ebel, Patrick and Meraner, Andrea and Schmitt, Michael and Zhu, Xiao Xiang},
  journal   = TGRS,
  year      = {2021},
  publisher = {IEEE},
  volume    = {59},
  number    = {7},
  pages     = {5866--5878}
}

@article{CUHK-CR,
  author  = {Sui, Jialu and Ma, Yiyang and Yang, Wenhan and Zhang, Xiaokang and Pun, Man-On and Liu, Jiaying},
  journal = TGRS,
  title   = {Diffusion Enhancement for Cloud Removal in Ultra-Resolution Remote Sensing Imagery},
  year    = {2024},
  volume  = {62},
  pages   = {1--14}
}

@article{CIALGC,
  title   = {Assessing the accuracy of blending Landsat--MODIS surface reflectances in two landscapes with contrasting spatial and temporal dynamics: A framework for algorithm selection},
  journal = RSE,
  volume  = {133},
  pages   = {193--209},
  year    = {2013},
  issn    = {0034-4257},
  author  = {Irina V. Emelyanova and Tim R. McVicar and Thomas G. {Van Niel} and Ling Tao Li and Albert I.J.M. {van Dijk}}
}

@article{zhang2022dense,
  title     = {Dense haze removal based on dynamic collaborative inference learning for remote sensing images},
  author    = {Zhang, Libao and Wang, Shan},
  journal   = TGRS,
  volume    = {60},
  pages     = {1--16},
  year      = {2022},
  publisher = {IEEE}
}

@article{RRSHID,
  author  = {Zhu, Zeng-Hui and Lu, Wei and Chen, Si-Bao and Ding, Chris H. Q. and Tang, Jin and Luo, Bin},
  journal = TGRS,
  title   = {Real-World Remote Sensing Image Dehazing: Benchmark and Baseline},
  year    = {2025},
  volume  = {63},
  pages   = {1--14}
}

@article{RSID,
  title     = {Trinity-{N}et: Gradient-guided Swin transformer-based remote sensing image dehazing and beyond},
  author    = {Chi, Kaichen and Yuan, Yuan and Wang, Qi},
  journal   = TGRS,
  volume    = {61},
  pages     = {1--14},
  year      = {2023},
  publisher = {IEEE}
}

@article{PanCollection,
  title     = {Machine learning in pansharpening: A benchmark, from shallow to deep networks},
  author    = {Deng, Liang-Jian and Vivone, Gemine and Paoletti, Mercedes E and Scarpa, Giuseppe and He, Jiang and Zhang, Yongjun and Chanussot, Jocelyn and Plaza, Antonio},
  journal   = GRSM,
  volume    = {10},
  number    = {3},
  pages     = {279--315},
  year      = {2022},
  publisher = {IEEE}
}

@article{Sen2Venus,
  title     = {Sen2ven$\mu$s, a dataset for the training of sentinel-2 super-resolution algorithms},
  author    = {Michel, Julien and Vinasco-Salinas, Juan and Inglada, Jordi and Hagolle, Olivier},
  journal   = {Data},
  volume    = {7},
  number    = {7},
  pages     = {96},
  year      = {2022},
  publisher = {MDPI}
}

@article{SARDespecklingFilter,
  title     = {Labeled dataset for training despeckling filters for {SAR} imagery},
  author    = {V{\'a}squez-Salazar, Rub{\'e}n Dar{\'\i}o and Cardona-Mesa, Ahmed Alejandro and G{\'o}mez, Luis and Travieso-Gonz{\'a}lez, Carlos M and Garavito-Gonz{\'a}lez, Andr{\'e}s F and V{\'a}squez-Cano, Esteban},
  journal   = DIB,
  volume    = {53},
  pages     = {110065},
  year      = {2024},
  publisher = {Elsevier}
}
}

\ifarxiv \clearpage \appendix 
\section{MoRA and softmax mixture approximation}
\label{sec:supp-mora}

This section gives the full tensor definitions behind the compact MoT/MoRA update in the main paper. With routing weights $\boldsymbol{\alpha}$ (components $\alpha_e$), expert value projections are fused and low-rank adapters supply query/key increments:
\begin{equation}
\begin{aligned}
\mathbf{V} &= \mathbf{F} \sum_{e \in \mathcal{K}} \alpha_e \mathbf{W}_{V,e}, \\
\Delta \mathbf{Q} &= \sum_{e \in \mathcal{K}} \alpha_e \mathbf{F}\,\mathbf{A}_{Q,e}\,\mathbf{B}_{Q,e}^{\top}, \\
\Delta \mathbf{K} &= \sum_{e \in \mathcal{K}} \alpha_e \mathbf{F}\,\mathbf{A}_{K,e}\,\mathbf{B}_{K,e}^{\top}.
\end{aligned}
\label{eq:supp-mot-fusion}
\end{equation}
With $\mathbf{Q}_0 = \mathbf{F}\mathbf{W}_Q$ and $\mathbf{K}_0 = \mathbf{F}\mathbf{W}_K$, the single-pass output in the main text is
\begin{equation}
\mathbf{Y}_{\text{MoT}} = \mathrm{Softmax}\left(\frac{(\mathbf{Q}_0 + \Delta \mathbf{Q})(\mathbf{K}_0 + \Delta \mathbf{K})^{\top}}{\sqrt{d_k}}\right) \mathbf{V}.
\label{eq:supp-mot-ymot}
\end{equation}

The forward pass~\eqref{eq:supp-mot-ymot} approximates the following \emph{exact} mixture of per-expert attentions:
\begin{equation}
\mathbf{Y}_{\text{multi}} = \sum_{e \in \mathcal{K}} \alpha_e \cdot \mathrm{Softmax}\left(\frac{(\mathbf{Q}_0 + \Delta \mathbf{Q}_e)(\mathbf{K}_0 + \Delta \mathbf{K}_e)^{\top}}{\sqrt{d_k}}\right) \mathbf{V}_e,
\label{eq:supp-mot-exact}
\end{equation}
where $\Delta \mathbf{Q}_e = \mathbf{F}\mathbf{A}_{Q,e}\mathbf{B}_{Q,e}^{\top}$, $\Delta \mathbf{K}_e = \mathbf{F}\mathbf{A}_{K,e}\mathbf{B}_{K,e}^{\top}$, and $\mathbf{V}_e = \mathbf{F}\mathbf{W}_{V,e}$.

Let $\mathbf{A}_0 = \mathbf{Q}_0 \mathbf{K}_0^{\top} / \sqrt{d_k}$ and let $\Delta \mathbf{A}_e$ denote the attention logit increment induced by expert $e$. A first-order Taylor expansion of softmax at $\mathbf{A}_0$ yields
\begin{equation}
\mathrm{Softmax}(\mathbf{A}_0 + \Delta \mathbf{A}_e) = \mathrm{Softmax}(\mathbf{A}_0) + \mathbf{J}(\mathbf{A}_0)\, \Delta \mathbf{A}_e + \mathbf{R}_e,
\label{eq:supp-softmax-taylor}
\end{equation}
where $\mathbf{J}(\mathbf{A}_0)$ is the Jacobian of softmax at $\mathbf{A}_0$ and $\mathbf{R}_e$ collects higher-order remainder terms. Because the first-order term $\mathbf{J}(\mathbf{A}_0)\,\Delta\mathbf{A}_e$ is linear in $\Delta\mathbf{A}_e$,
\begin{equation}
\sum_{e \in \mathcal{K}} \alpha_e\, \mathrm{Softmax}(\mathbf{A}_0 + \Delta \mathbf{A}_e) \approx \mathrm{Softmax}\!\left(\mathbf{A}_0 + \sum_{e \in \mathcal{K}} \alpha_e\, \Delta \mathbf{A}_e\right).
\label{eq:supp-taylor-aggregation}
\end{equation}
The mismatch between the two sides is captured by aggregated remainders; schematically, $\mathbf{R}_{\text{diff}} = \sum_e \alpha_e \mathbf{R}_e - \mathbf{R}$, where $\mathbf{R}$ is the remainder of the single fused softmax path. LoRA initialization (one factor random, the other zero) gives $\Delta \mathbf{Q}_e = \Delta \mathbf{K}_e = \mathbf{0}$ at initialization, matching the small-perturbation regime; low rank and sparse routing limit perturbation growth during training. A single MoT pass thus cuts cost by roughly $|\mathcal{K}|$ relative to full per-expert attention while preserving diversity.

\section{LLaRS1M dataset details}
\label{sec:supp-llars1m}

\noindent\textbf{Sources.}
LLaRS1M merges public paired imagery for cloud removal, dehazing, pansharpening, super-resolution, SAR despeckling, and spatiotemporal fusion across optical multispectral, panchromatic, and SAR modalities. \emph{Cloud removal} combines SEN12MS-CR~\citep{SEN12MS-CR} (Sentinel-2/Sentinel-1 triplets worldwide), CUHK-CR~\citep{CUHK-CR} (Jilin-1 thin/thick cloud splits), and RICE~\citep{RICE} (cloudy/clear pairs with optional masks). \emph{Dehazing} mixes SateHaze1k~\citep{WACV2020_Haze1K} (semi-synthetic GF-2/GF-3 RGB--SAR triplets), RRSHID~\citep{RRSHID} (real haze--clear pairs clustered by thickness), RSID~\citep{RSID} (non-uniform haze biased to ships and airports), and DHID/LHID~\citep{zhang2022dense} (physically inspired synthesis from Google Earth and MVAID). \emph{Pansharpening} uses PansharpRSData~\citep{9082183} (IKONOS, QuickBird, GF-1, WorldView-2/3/4) and PanCollection~\citep{PanCollection} (WorldView-3, QuickBird, GF-2 with Wald-protocol test splits). \emph{Super-resolution} pairs Landsat-8 OLI with Sentinel-2 MSI in OLI2MSI~\citep{OLI2MSI} and aligns Sentinel-2 with VEN$\mu$S in SEN2VEN$\mu$S~\citep{Sen2Venus}. \emph{SAR despeckling} draws from SAR-despeckle-Dataset~\citep{SARDespeckleDataset} (multitemporal Sentinel-1 fusion) and SAR-despeckling-filter~\citep{SARDespecklingFilter} (Toronto VV stacks). \emph{Spatiotemporal fusion} covers CIA and LGC~\citep{CIALGC} (Landsat--MODIS over irrigated fields and heterogeneous forest/pasture) and AHB, TJ, and DX~\citep{li2020spatio} (Inner Mongolia phenology, Tianjin urban--crop mix, and Beijing Daxing airport growth). Together these sources span dense urban, cropland, forest, mountain, coast, and desert across climates and seasons. \cref{tab:supp-llars1m-basic,tab:supp-llars1m-pixel} summarize per-source basic information and pixel statistics, respectively.

\begin{table*}[t]
    \centering
    \footnotesize
    \setlength{\tabcolsep}{4pt}
    \renewcommand{\arraystretch}{1.2}
    \caption{Basic information of LLaRS1M sources.}
    \label{tab:supp-llars1m-basic}
    % \resizebox{\textwidth}{!}{%
    \begin{tabular}{llllllcr}
\toprule
Task & Dataset & Source & Input & GSD & Size & Bands & Samples \\
\midrule
\multirow{2}{*}{SR} & OLI2MSI\citep{OLI2MSI} & Landsat-8 / Sentinel-2 & MS & 30\,m / 10\,m & $160\times160$ & 3 & 5{,}325 \\
 & SEN2VEN$\mu$S\citep{Sen2Venus} & Sentinel-2 / VEN$\mu$S & MS & 10--20\,m / 5\,m & $128\times128$ / $256\times256$ & 4 & 132{,}955 \\
\midrule
\multirow{4}{*}{Pan} & \multirow{2}{*}{PansharpRSData\citep{9082183}} & \multirow{2}{*}{GF-1 / IKONOS / QB / WV-2/3/4} & MS & 1--4\,m & $256\times256$ & 4/8 & \multirow{2}{*}{2{,}270} \\
 &  &  & PAN & 0.3--1\,m & $1024\times1024$ & 1 &  \\
 & \multirow{2}{*}{PanCollection\citep{PanCollection}} & \multirow{2}{*}{GF-2 / QB / WV-2/3} & PAN & 0.3--0.8\,m & $64\times64$ / $256\times256$ & 1 & \multirow{2}{*}{51{,}928} \\
 &  &  & MS & 1.2--3.2\,m & $64\times64$ / $256\times256$ & 4/8 &  \\
\midrule
\multirow{4}{*}{Cloud} & CUHK-CR\citep{CUHK-CR} & Jilin-1 & RGB & 0.5\,m & $512\times512$ & 3 & 1{,}227 \\
 & RICE\citep{RICE} & -- & RGB & -- & $512\times512$ & 3 & 1{,}236 \\
 & \multirow{2}{*}{SEN12MS-CR\citep{SEN12MS-CR}} & \multirow{2}{*}{Sentinel-2 / Sentinel-1} & MS & 10\,m & $256\times256$ & 13 & \multirow{2}{*}{122{,}218} \\
 &  &  & SAR & 10\,m & $256\times256$ & 2 &  \\
\midrule
\multirow{5}{*}{Haze} & SateHaze1k\citep{WACV2020_Haze1K} & GF-2 / GF-3 & RGB & $\sim$1\,m & $512\times512$ & 3 & 1{,}200 \\
 & RRSHID\citep{RRSHID} & GF PMS & RGB & $\leq$2\,m & $256\times256$ & 3 & 3{,}053 \\
 & RSID\citep{RSID} & -- & RGB & -- & $256\times256$ & 3 & 1{,}000 \\
 & DHID\citep{zhang2022dense} & Google Earth / MVAID & RGB & 0.2--153\,m & $512\times512$ & 3 & 14{,}990 \\
 & LHID\citep{zhang2022dense} & Google Earth & RGB & 0.2--153\,m & $512\times512$ & 3 & 31{,}517 \\
\midrule
\multirow{10}{*}{STF} & \multirow{2}{*}{CIA\citep{CIALGC}} & \multirow{2}{*}{Landsat / MODIS} & Landsat MS & 30\,m & $1280\times1792$ & 6 & \multirow{2}{*}{21} \\
 &  &  & MODIS MS & 500\,m & $1280\times1792$ & 6 &  \\
 & \multirow{2}{*}{AHB\citep{li2020spatio}} & \multirow{2}{*}{Landsat-8 / MODIS} & Landsat MS & 30\,m & $2800\times2480$ & 6 & \multirow{2}{*}{33} \\
 &  &  & MODIS MS & 500\,m & $2800\times2480$ & 6 &  \\
 & \multirow{2}{*}{DX\citep{li2020spatio}} & \multirow{2}{*}{Landsat-8 / MODIS} & Landsat MS & 30\,m & $1640\times1640$ & 6 & \multirow{2}{*}{37} \\
 &  &  & MODIS MS & 500\,m & $1640\times1640$ & 6 &  \\
 & \multirow{2}{*}{LGC\citep{CIALGC}} & \multirow{2}{*}{Landsat / MODIS} & Landsat MS & 30\,m & $3072\times2560$ & 6 & \multirow{2}{*}{17} \\
 &  &  & MODIS MS & 500\,m & $3072\times2560$ & 6 &  \\
 & \multirow{2}{*}{TJ\citep{li2020spatio}} & \multirow{2}{*}{Landsat-8 / MODIS} & Landsat MS & 30\,m & $1970\times2100$ & 6 & \multirow{2}{*}{34} \\
 &  &  & MODIS MS & 500\,m & $1970\times2100$ & 6 &  \\
\midrule
\multirow{2}{*}{SAR} & SAR-despeckling-filter\citep{SARDespecklingFilter} & Sentinel-1 & SAR & 10\,m & $512\times512$ & 1 & 1{,}600 \\
 & SAR-despeckle-Dataset\citep{SARDespeckleDataset} & Sentinel-1 & SAR & 10\,m & $256\times256$ & 1 & 936 \\
\bottomrule
\end{tabular}%
% }
\end{table*}

\begin{table*}[t]
    \centering
    \footnotesize
    \setlength{\tabcolsep}{4pt}
    \renewcommand{\arraystretch}{1.2}
    \caption{Pixel statistics of LLaRS1M sources.}
    \label{tab:supp-llars1m-pixel}
    % \resizebox{\textwidth}{!}{%
    \begin{tabular}{lllllrrrr}
\toprule
Task & Dataset & Modality & Format & dtype & Mean & Std & Min & Max \\
\midrule
\multirow{2}{*}{SR} & OLI2MSI\citep{OLI2MSI} & MS & TIFF & float32 & 0.0736 & 0.0150 & 0.0336 & 0.1492 \\
 & SEN2VEN$\mu$S\citep{Sen2Venus} & MS & PT & int16 & 1049.70 & 398.02 & 91.57 & 4405.95 \\
\midrule
\multirow{4}{*}{Pan} & \multirow{2}{*}{PansharpRSData\citep{9082183}} & MS & MAT & uint16 & 346.48 & 163.35 & 106.80 & 1127.70 \\
 &  & PAN & MAT & uint16 & 335.57 & 159.21 & 145.75 & 1099.51 \\
 & \multirow{2}{*}{PanCollection\citep{PanCollection}} & PAN & H5 & float64 & 295.78 & 149.00 & 124.87 & 951.15 \\
 &  & MS & H5 & float64 & 308.43 & 129.09 & 67.73 & 935.95 \\
\midrule
\multirow{4}{*}{Cloud} & CUHK-CR\citep{CUHK-CR} & RGB & PNG & uint8 & 147.34 & 29.11 & 122.19 & 213.15 \\
 & RICE\citep{RICE} & RGB & PNG & uint8 & 113.43 & 43.85 & 81.55 & 159.20 \\
 & \multirow{2}{*}{SEN12MS-CR\citep{SEN12MS-CR}} & MS & TIFF & uint16 & 2258.98 & 1516.04 & 101.61 & 5961.58 \\
 &  & SAR & TIFF & float32 & $-14.77$ & 4.64 & $-29.66$ & $-2.03$ \\
\midrule
\multirow{5}{*}{Haze} & SateHaze1k\citep{WACV2020_Haze1K} & RGB & PNG & uint8 & 194.41 & 31.01 & 81.12 & 254.43 \\
 & RRSHID\citep{RRSHID} & RGB & PNG & uint8 & 156.20 & 21.54 & 119.37 & 224.32 \\
 & RSID\citep{RSID} & RGB & PNG & uint8 & 175.74 & 32.07 & 125.27 & 236.21 \\
 & DHID\citep{zhang2022dense} & RGB & JPG & uint8 & 192.61 & 26.50 & 168.96 & 221.09 \\
 & LHID\citep{zhang2022dense} & RGB & JPG & uint8 & 151.64 & 27.79 & 105.86 & 208.19 \\
\midrule
\multirow{10}{*}{STF} & \multirow{2}{*}{CIA\citep{CIALGC}} & Landsat MS & TIFF & int16 & 1648.58 & 644.61 & 140.55 & 4818.55 \\
 &  & MODIS MS & TIFF & int16 & 1723.64 & 499.21 & 338.59 & 3928.09 \\
 & \multirow{2}{*}{AHB\citep{li2020spatio}} & Landsat MS & TIFF & uint8 & 41.97 & 16.35 & 10.18 & 101.65 \\
 &  & MODIS MS & TIFF & uint8 & 52.10 & 13.47 & 13.29 & 104.47 \\
 & \multirow{2}{*}{DX\citep{li2020spatio}} & Landsat MS & TIFF & uint8 & 44.70 & 18.57 & 2.79 & 166.37 \\
 &  & MODIS MS & TIFF & uint8 & 45.51 & 13.77 & 12.95 & 92.76 \\
 & \multirow{2}{*}{LGC\citep{CIALGC}} & Landsat MS & TIFF & int16 & 1404.21 & 536.49 & 180.78 & 4557.67 \\
 &  & MODIS MS & TIFF & int16 & 1340.16 & 457.55 & 265.06 & 3015.61 \\
 & \multirow{2}{*}{TJ\citep{li2020spatio}} & Landsat MS & TIFF & uint8 & 47.38 & 27.19 & 2.61 & 220.17 \\
 &  & MODIS MS & TIFF & uint8 & 46.66 & 21.42 & 13.42 & 93.94 \\
\midrule
\multirow{2}{*}{SAR} & SAR-despeckling-filter\citep{SARDespecklingFilter} & SAR & TIFF & uint8 & 81.50 & 50.04 & 22.76 & 189.46 \\
 & SAR-despeckle-Dataset\citep{SARDespeckleDataset} & SAR & MAT & float32 & 2.03e+05 & 2.77e+06 & 809.22 & 8.34e+07 \\
\bottomrule
\end{tabular}%
% }
\end{table*}

\noindent\textbf{Simulated degradations.}
Clean references for the six synthetic families are drawn uniformly from the training splits of the real paired sources above. Six tasks use on-the-fly corruption of high-quality references: deblurring (Gaussian, motion, mean, and disk-out-of-focus kernels with random size and strength), denoising (mixtures of Gaussian, uniform, Poisson, Rayleigh, gamma, salt-and-pepper, impulse, and multiplicative speckle noise), destriping (periodic, non-uniform, and rotated stripe patterns), global brightness rescaling with random gain, histogram equalization and CLAHE-style mappings as synthetic inputs, and 2--98\% per-channel linear stretch. Each family samples severity at random per step for robustness.

\noindent\textbf{Unification.}
Raw tiles arrive in heterogeneous formats, bit depths, and channel counts. Loader pipelines cast tensors to floating point, apply dataset-wise percentile clipping and min--max scaling (or division by 255 for 8-bit data), random-crop to the training patch size when possible, repeat or subset channels to the model width, and emit tuples $(\tilde{\mathbf{x}}, \tilde{\mathbf{y}}, \text{task id}, \text{prompt})$ with spatially aligned degraded and clean patches.

\noindent\textbf{Prompts.}
Natural-language instructions are generated with multiple commercial large language models under task-specific system prompts; paraphrases and style variants are deduplicated and pooled per task so each minibatch samples a diverse instruction.

\section{Additional Experimental Results}
\label{sec:supp-additional-results}

\subsection{Training and Evaluation Setup}

Training uses two NVIDIA A100 80\,GB GPUs, Python~3.12, and PyTorch~2.6. We take a stratified 5\% subset of LLaRS1M so all eleven tasks stay balanced. Optimization is AdamW with base learning rate $10^{-4}$, weight decay $0.01$, batch size 48, and 300 epochs. Patches are $128\times128$ with random flips; each sample draws one prompt from its task family. Spatiotemporal fusion uses oversampling to balance counts. The text tower is DistilBERT~\citep{DistilBERT} (768-D); the image tower is ResNet-18~\citep{ResNet} features projected to 768-D. Unified channel dimension $C=20$, alignment slots $S=32$, Sinkhorn iterations $T=8$. The U-Net~\citep{UNet} follows Restormer~\citep{Restormer} blocks with base width 32, stage depths $[2,4,6,8]$, two refinement blocks, heads $[1,2,4,8]$, and FFN expansion 2.66. Every MoE stage uses eight experts with Top-$2$ activation. Step-level DWA uses $\gamma=0.7$ and temperature $0.1$. At test time prompts are fixed. Simulated degradations match training: Gaussian noise $\sigma=0.05$; $5\times5$ motion blur at $45^\circ$; synthetic destriping with seeded parameters; 2--98\% linear stretch reversal; standard histogram equalization reversal; brightness targets 1.2$\times$ or 0.7$\times$ inputs.

Baselines are MPRNet~\citep{MPRNet}, Restormer~\citep{Restormer}, GridFormer~\citep{GridFormer}, PromptIR~\citep{PromptIR}, AMIR~\citep{AMIR}, HOGformer~\citep{HOGformer}, and MoCE-IR~\citep{MoCEIR}, all multitask-trained on the same subset with identical dataloading and evaluation code. We report PSNR and SSIM for pixel fidelity, SAM and ERGAS for spectral and radiometric fidelity. Higher PSNR/SSIM and lower SAM/ERGAS are better.

\subsection{Dataset Visualization}

\begin{figure}[h]
    \centering
    \includegraphics[width=\linewidth]{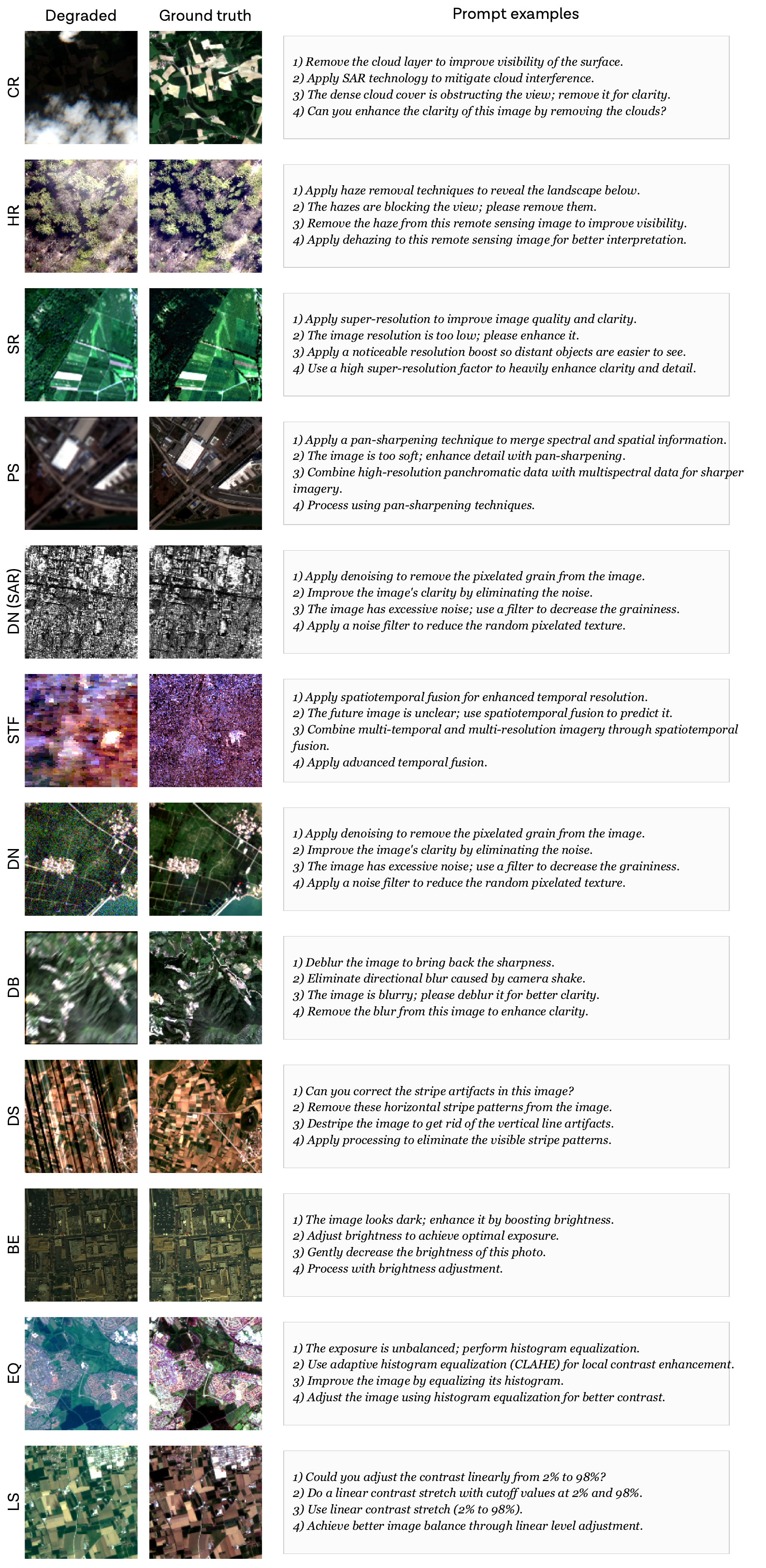}
    \caption{LLaRS1M examples.}
    \label{fig:llars1m-samples}
\end{figure}

\cref{fig:llars1m-samples} stresses visible diversity: cloud occlusion, haze veils, blur and speckle-like noise, periodic stripes, globally dimmed radiometry, low-resolution optical inputs, multisensor pansharpening pairs, and multi-date stacks for fusion. The montage underscores that tuples pair heterogeneous sensors and degradation physics, giving multi-task training a wide visual and radiometric span.

\subsection{Test-Time Prompts}

\begin{table}[h]
    \caption{Fixed text prompts used at test time for each task.}
    \label{tab:test-prompts}
    \centering
    \footnotesize
    \renewcommand{\arraystretch}{1.12}
    \resizebox{1\linewidth}{!}{
    \begin{tabular}{p{0.05\linewidth}p{0.8\linewidth}}
    \toprule
    Task & Prompt \\
    \midrule
    CR & \textit{Clear the clouds from this satellite image to reveal the ground.} \\
    SR & \textit{Increase the image's resolution for a more detailed view.} \\
    PS & \textit{Combine panchromatic and multispectral images to produce a higher-resolution color image.} \\
    HR & \textit{Dehaze this image to reveal the underlying surface features.} \\
    DN & \textit{Enhance the image clarity by reducing the noise level.} \\
    DB & \textit{Deblur the image to recover fine structure and detail.} \\
    DS & \textit{Remove the banding effect to improve image appearance.} \\
    EQ & \textit{Bring out details in shadows and highlights by equalizing the image's histogram.} \\
    LS & \textit{Balance the image by adjusting levels linearly.} \\
    BE & \textit{The image is a bit dim; increase brightness slightly.} \\
    STF & \textit{Predict a high-resolution image at the target date using spatiotemporal fusion.} \\
    \bottomrule
    \end{tabular}
    }
\end{table}

\subsection{Ablation Study Details}

\subsubsection{Conditioning Mechanisms}

\begin{figure}[htbp]
    \centering
    \includegraphics[width=\linewidth]{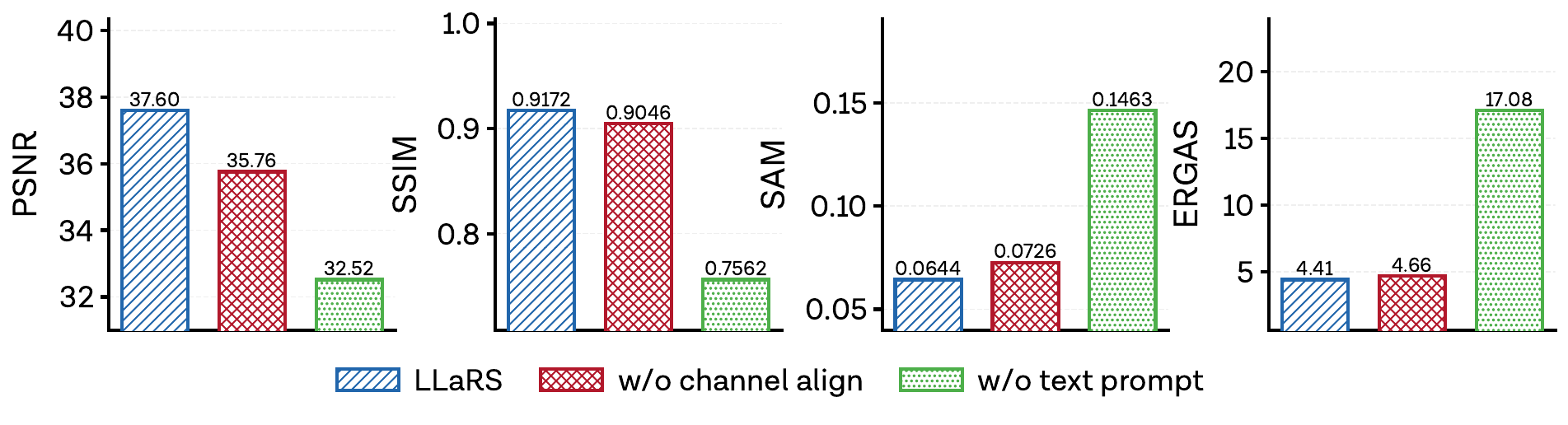}
    \caption{Contribution analysis of text prompt and OT-based channel alignment.}
    \label{fig:ablation-binary}
\end{figure}

\cref{fig:ablation-binary} compares the full model against variants that remove either text prompts or OT-based channel alignment. Removing text prompts causes the most severe degradation across all metrics, because the routing network loses explicit task identity and must infer the intended operation solely from degraded image statistics. This ambiguity particularly hurts tasks with similar visual characteristics, such as dehazing versus cloud removal, or different noise types. Without textual guidance, the gating network struggles to consistently activate the appropriate experts, leading to inconsistent routing and reduced specialization. Removing band matching while retaining text prompts shows a different failure mode: PSNR and SSIM remain relatively stable, but SAM and ERGAS increase substantially. This pattern indicates that naive zero-padding or channel replication fails to preserve the semantic relationships between spectral bands. The entropic OT formulation in this alignment stage~\citep{sinkhorn1967concerning,cuturi2013sinkhorn} explicitly matches input channels to the model's internal representation based on statistical similarity, ensuring that spectral correlations are maintained even when processing inputs with heterogeneous channel counts. Together, text prompts and the transport-based front-end address orthogonal challenges in unified multitask learning: task disambiguation and input standardization.

\subsubsection{Number of Experts}

\begin{table}[t]
    \caption{Effect of the number of experts $E$ on model performance.}
    \label{tab:ablation-num-experts}
    \centering
    \footnotesize
    \renewcommand{\arraystretch}{1.15}
    \resizebox{1\linewidth}{!}{
    \begin{tabular}{ccccccc}
    \toprule
    $E$ & PSNR$\uparrow$ & SSIM$\uparrow$ & SAM$\downarrow$ & ERGAS$\downarrow$ & Params(M) & FLOPs(G) \\
    \midrule
    $0$ & 32.76 & 0.8371 & 0.1399 & 14.6733 & 11.85 & 96.1 \\
    $2$ & 34.04 & 0.8894 & 0.0853 & 6.2959 & 42.31 & 109.2 \\
    $4$ & 35.00 & 0.8991 & 0.0763 & 4.9168 & 56.08 & 109.3 \\
    $8$ & \textbf{37.60} & \textbf{0.9172} & \textbf{0.0644} & \textbf{4.4069} & 83.64 & 109.6 \\
    $16$ & 35.75 & 0.9001 & 0.0752 & 7.2115 & 138.74 & 110.1 \\
    \bottomrule
    \end{tabular}
    }
\end{table}

\cref{tab:ablation-num-experts} shows that performance improves steadily as we increase from zero experts to eight experts per MoE layer, but degrades when expanding to sixteen experts. This degradation likely stems from insufficient training data to fully utilize all experts, leading to redundant or underutilized pathways. While the parameter count grows roughly linearly with expert count, the fused implementation keeps FLOPs growth modest, making eight experts a practical choice that balances quality and efficiency.

\subsubsection{Multitask Optimization Strategies}

\begin{table}[t]
    \caption{Comparison of multi-task optimization strategies.}
    \label{tab:ablation-dwa}
    \centering
    \footnotesize
    \renewcommand{\arraystretch}{1.15}
    \resizebox{1\linewidth}{!}{
    \begin{tabular}{lcccccc}
    \toprule
    Strategy & PSNR$\uparrow$ & SSIM$\uparrow$ & SAM$\downarrow$ & ERGAS$\downarrow$ & Time/step (ms) & Extra mem \\
    \midrule
    Equal wgt & 34.31 & 0.8696 & 0.1000 & 13.0818 & 542 & -- \\
    PCGrad~\citep{PCGrad} & -- & -- & -- & -- & 2884 & +18.7 GB \\
    MGDA~\citep{MGDA} & -- & -- & -- & -- & 2661 & +21.4 GB \\
    CAGrad~\citep{CAGrad} & -- & -- & -- & -- & 2773 & +23.8 GB \\
    NashMTL~\citep{NashMTL} & -- & -- & -- & -- & 2798 & +13.9 GB \\
    GradNorm~\citep{GradNorm} & 34.56 & 0.8338 & 0.0997 & 12.0695 & 2841 & +29.0 GB \\
    \textbf{DWA} & \textbf{37.60} & \textbf{0.9172} & \textbf{0.0644} & \textbf{4.4069} & 536 & -- \\
    \bottomrule
    \end{tabular}
    }
\end{table}

\cref{tab:ablation-dwa} compares our step-level Dynamic Weight Adjustment (DWA) against equal weighting and gradient-based methods. Equal weighting underperforms because it cannot adapt to tasks that learn at different speeds, allowing fast-converging tasks to dominate the gradient signal. Gradient manipulation methods like PCGrad, CAGrad, and NashMTL offer theoretical advantages but require one backward pass per task, multiplying both computation time and memory consumption by the number of tasks. With eleven tasks, this overhead becomes prohibitive, and NashMTL additionally solves a quadratic program at each step. GradNorm provides a middle ground by dynamically adjusting task weights based on gradient norms, but requires tracking additional statistics and introduces extra hyperparameters. Our DWA approach achieves comparable wall-clock time to equal weighting while improving all four metrics, because task weights are computed in a forward pass from loss ratios and exponential moving averages without requiring gradient computation or backpropagation through the weighting mechanism.

\subsection{Model Efficiency Comparison}

\begin{table}[t]
    \caption{Comparison of model efficiency.}
    \label{tab:efficiency}
    \centering
    \footnotesize
    \renewcommand{\arraystretch}{1.15}
    \resizebox{1\linewidth}{!}{
    \begin{tabular}{lcccc}
    \toprule
    Method & Params(M) & FLOPs(G) & Time (ms) & Avg.PSNR$\uparrow$ \\
    \midrule
    MPRNet~\citep{MPRNet} & 3.67 & 285.9 & 45 & 32.49 \\
    Restormer~\citep{Restormer} & 26.15 & 312.7 & 73 & 36.07 \\
    GridFormer~\citep{GridFormer} & 34.18 & 742.1 & 390 & 35.24 \\
    PromptIR~\citep{PromptIR} & 35.61 & 348.3 & 77 & 36.25 \\
    AMIR~\citep{AMIR} & 23.57 & 281.5 & 96 & 34.15 \\
    HOGformer~\citep{HOGformer} & 30.00 & 426.9 & 308 & 35.33 \\
    MoCE-IR~\citep{MoCEIR} & 23.08 & 186.2 & 133 & 35.31 \\
    \textbf{LLaRS} & 83.64 & 109.6 & 386 & \textbf{37.60} \\
    \bottomrule
    \end{tabular}
    }
\end{table}

\cref{tab:efficiency} compares parameter counts, theoretical FLOPs at $256 \times 256$ resolution, measured latency on a single A100 GPU, and average PSNR across all tasks. Our model contains more parameters than Restormer due to the multiple expert pathways, but achieves lower FLOPs because the fused expert implementation avoids computing full dense attention repeatedly for each expert. The Top-$2$ gating mechanism activates only a subset of experts per input, further reducing computation compared to a naive implementation that would evaluate all experts. However, wall-clock latency remains higher than both Restormer and PromptIR despite the lower FLOP count, reflecting the gap between theoretical arithmetic operations and practical memory bandwidth constraints. The routing mechanism introduces additional memory traffic as activations flow through gating networks and expert selection logic, and the irregular computation pattern from dynamic expert selection prevents optimal GPU utilization. This latency overhead represents a trade-off between quality and speed: the model achieves substantially higher PSNR per parameter and per FLOP than competing methods, but requires more time per image due to architectural complexity.

\subsection{Fine-Tuning Parameter Statistics}

\begin{table}[t]
    \caption{Parameter statistics of different fine-tuning methods for each model.}
    \label{tab:finetune-params}
    \centering
    \footnotesize
    \setlength{\tabcolsep}{4pt}
    \renewcommand{\arraystretch}{1.2}
    \resizebox{0.9\linewidth}{!}{
    \begin{tabular}{ll rrr}
    \toprule
    Model & FT method & Total params & Train params & Train.Ratio \\
    \midrule
    \multirow{6}{*}{PromptIR}
    & Full & 35.61M & 35.61M & 100.00\% \\
    & LoRA & 36.88M & 1.27M & 3.44\% \\
    & Adapter & 56.90M & 21.28M & 37.41\% \\
    & DoRA & 36.97M & 1.36M & 3.68\% \\
    & SSF & 35.94M & 329.7K & 0.92\% \\
    & BitFit & 35.61M & 17.4K & 0.05\% \\
    \midrule
    \multirow{6}{*}{Restormer}
    & Full & 26.15M & 26.15M & 100.00\% \\
    & LoRA & 27.21M & 1.06M & 3.89\% \\
    & Adapter & 44.40M & 18.25M & 41.11\% \\
    & DoRA & 27.28M & 1.14M & 4.16\% \\
    & SSF & 26.43M & 282.7K & 1.07\% \\
    & BitFit & 26.15M & 15.0K & 0.06\% \\
    \midrule
    \multirow{6}{*}{MoCE-IR}
    & Full & 23.08M & 23.08M & 100.00\% \\
    & LoRA & 24.21M & 1.13M & 4.69\% \\
    & Adapter & 39.20M & 16.12M & 41.13\% \\
    & DoRA & 24.29M & 1.21M & 4.98\% \\
    & SSF & 23.33M & 249.3K & 1.07\% \\
    & BitFit & 23.08M & 99.9K & 0.43\% \\
    \midrule
    \multirow{6}{*}{\textbf{LLaRS}}
    & Full & 83.64M & 83.64M & 100.00\% \\
    & LoRA & 87.44M & 3.80M & 4.35\% \\
    & Adapter & 95.58M & 11.94M & 12.50\% \\
    & DoRA & 87.52M & 3.89M & 4.44\% \\
    & SSF & 83.82M & 184.1K & 0.22\% \\
    & BitFit & 83.64M & 71.1K & 0.09\% \\
    \bottomrule
    \end{tabular}
    }
\end{table}

\cref{tab:finetune-params} lists total parameters, trainable parameters, and trainable ratios for each model and method.

\subsection{Qualitative Comparison Results}
\label{sec:supp-qualitative-results}

The following figures provide more comprehensive qualitative comparisons.

\begin{figure}[htbp]
    \centering
    \includegraphics[width=\linewidth]{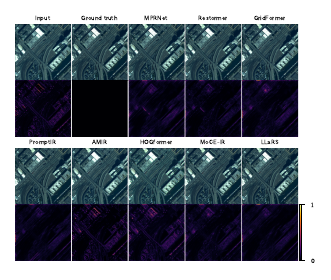}
    \caption{Model predictions and error maps for deblurring.}
    \label{fig:qualitative-deblur}
\end{figure}

\begin{figure}[htbp]
    \centering
    \includegraphics[width=\linewidth]{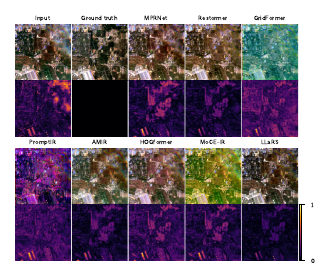}
    \caption{Model predictions and error maps for cloud removal.}
    \label{fig:qualitative-cloud}
\end{figure}

\begin{figure}[htbp]
    \centering
    \includegraphics[width=\linewidth]{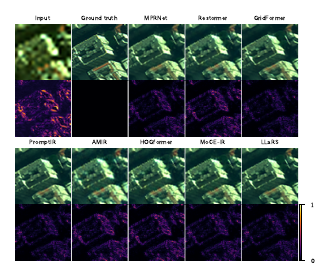}
    \caption{Model predictions and error maps for pansharpening.}
    \label{fig:qualitative-pan}
\end{figure}

\begin{figure}[htbp]
    \centering
    \includegraphics[width=\linewidth]{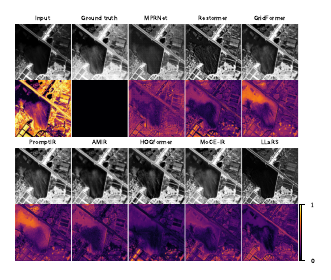}
    \caption{Model predictions and error maps for histogram equalization.}
    \label{fig:qualitative-histeq}
\end{figure}

\begin{figure}[htbp]
    \centering
    \includegraphics[width=\linewidth]{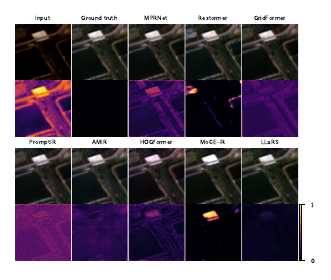}
    \caption{Model predictions and error maps for brightness enhancement.}
    \label{fig:qualitative-bright}
\end{figure}

\begin{figure*}[t]
    \centering
    \includegraphics[width=\linewidth]{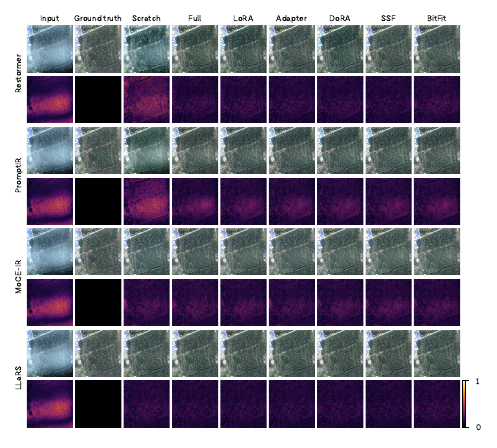}
    \caption{Fine-tuning qualitative comparison for dehazing.}
    \label{fig:finetune-qual-dehaze}
\end{figure*}

\begin{figure*}[t]
    \centering
    \includegraphics[width=\linewidth]{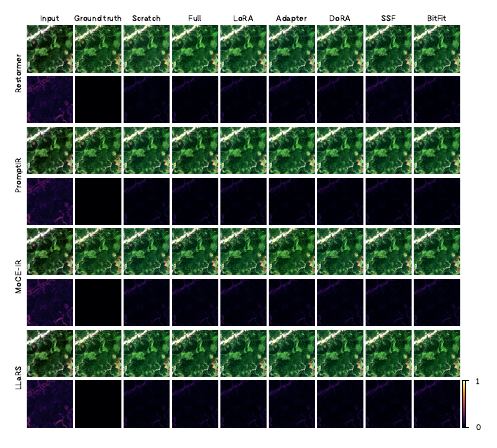}
    \caption{Fine-tuning qualitative comparison for super-resolution.}
    \label{fig:finetune-qual-superres}
\end{figure*}

\begin{figure*}[t]
    \centering
    \includegraphics[width=\linewidth]{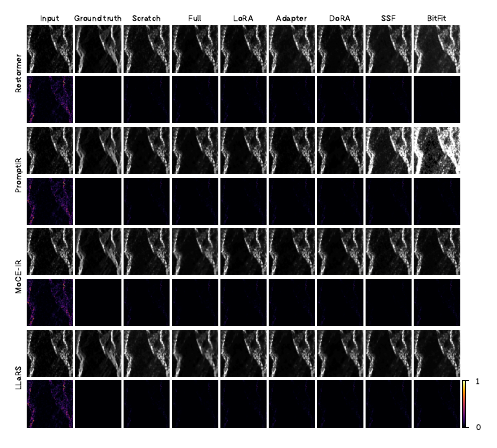}
    \caption{Fine-tuning qualitative comparison for SAR despeckling.}
    \label{fig:finetune-qual-sar-speckle}
\end{figure*}

 \fi

\end{document}